# Neuromorphic scaling advantages for energy-efficient random walk computations

J. Darby Smith, Aaron J. Hill, Leah E. Reeder, Brian C. Franke, Richard B. Lehoucq, Ojas Parekh, William Severa, James B. Aimone*

Neural Exploration & Research Laboratory

Sandia National Laboratories

Albuquerque, NM 87185

* To whom correspondence should be addressed: jbaimon@sandia.gov

**Computing stands to be radically improved by neuromorphic computing (NMC) approaches inspired by the brain's incredible efficiency and capabilities. Most NMC research, which aims to replicate the brain's computational structure and architecture in man-made hardware, has focused on artificial intelligence; however, less explored is whether this brain-inspired hardware can provide value beyond cognitive tasks. We demonstrate that high-degree parallelism and configurability of spiking neuromorphic architectures makes them well-suited to implement random walks via discrete time Markov chains. Such random walks are useful in Monte Carlo methods, which represent a fundamental computational tool for solving a wide range of numerical computing tasks. Additionally, we show how the mathematical basis for a probabilistic solution involving a class of stochastic differential equations can leverage those simulations to provide solutions for a range of broadly applicable computational tasks. Despite being in an early development stage, we find that NMC platforms, at a sufficient scale, can drastically reduce the energy demands of high-performance computing (HPC) platforms.**

The efficiency of biological nervous systems has intrigued even the earliest designers of computing systems [24, 39], but the theoretical value of neuromorphic hardware remains unclear. While quantum computing offers clear fundamental advantages at scale [36], the advantages of NMC are more subtle, a fact that has muted enthusiasm despite the increasing ability to develop large scale neural processors today [9, 13, 26]. Nonetheless, in addition to the advanced cognitive capabilities, there are several architectural features of most nervous systems that may yield advantages including the high degree of connectivity between neurons, the colocation of processing and memory, and the use of action potentials (i.e., spikes) to communicate.

Algorithms research for spiking neuromorphic hardware has primarily focused on its suitability for deep learning and other emerging AI algorithms [31, 35]. This application is straight-forward, given the alignment of neural architectures with neural networks, and it can be expected that the value of NMC will grow as AI algorithms derive further inspiration from the brain [1]. However, the impact of NMC beyond cognitive applications is less clear. Quantum computing provides a precedent for emerging hardware to have impact beyond its original inspiration: while quantum computing was conceived as a means for efficient chemistry simulations [11, 22], it is now recognized that it can impact a much broader range of computing applications [4, 18, 36]. Along these lines, there is growing evidence that neuromorphic hardware can provide theoretical complexity advantages on a growing set of non-cognitive, non-AI applications [2, 3, 7, 12, 25, 27-29, 33, 37]. Unlike quantum computing, which still faces technical challenges in scaling up to sizes necessary for real-world impact (as noted by the recent findings concerning *quantum supremacy* [4]), NMC platforms can already be scaled to non-trivial sizes, with several multi-chip spiking NMC systems achieving scales of over a hundred million neurons. Nevertheless, NMC systems remain smaller and less efficient than the human brain, and the critical scales for NMC remain unknown since the appropriateness of an analogous concept of *neuromorphic supremacy* remains unclear.

Identifying NMC's value for an application is complicated by the fact that its advantage primarily derives from its energy-efficiency as opposed to a promise of faster computation (although speed benefits remain a possibility, and because NMC is an immature technology compared to conventional von Neumann (VN) systems, which have been optimized and advanced over decades in both hardware and software. **We define an algorithm as having a *neuromorphic advantage* if that algorithm shows a demonstrable advantage (compared to a VN architecture) in one resource (e.g., energy) while exhibiting comparable or better scaling in other resources (e.g., time).** Given NMC's currently realized advantages in power consumption, we are seeking algorithms that show comparable or better time-scaling compared to a VN architecture while still requiring less total energy (i.e., "energy efficiency") to perform the same computation.

Observing a *neuromorphic advantage* for non-cognitive applications should not be taken as a given, as the specialization of computer architectures to improve performance on a subset of tasks (in the case of NMC, towards the brain) will result in degraded performance in other tasks [41]. Therefore, observing a neuromorphic advantage on non-cognitive applications would demonstrate that NMC can have a broader impact than previously assumed and provide a concrete framework by which to develop the technology.

In this paper, we identify for the first time an explicit neuromorphic advantage for large-scale spiking neuromorphic hardware on a fundamental numerical computing task: solving partial integro-differential equations (PIDEs) that have probabilistic representations involving a jump-diffusion stochastic differential equation (SDE). The solutions to these PIDEs can be approximated by averaging over many independent random walks (RWs), a process often referred to as Monte Carlo. Diffusion is a quintessential component of the underlying SDEs used in the probabilistic solution of the PIDEs. We can show our NMC algorithm for generating RW approximations to diffusion satisfies our neuromorphic advantage criteria on two current large-scale neuromorphic platforms: the IBM Neurosynaptic system [26], also known as TrueNorth and introduced in 2014, and the Intel Loihi system [9], introduced in 2018. While distinct neural architectures, both directly implement a large number of neurons in silicon (1 million and 128 thousand per chip, respectively), are readily-scalable to multi-chip platforms, and are reflective of the long-term technology trends in spiking neuromorphic hardware. We then show that our NMC algorithm for random walks can be extended to account for more sophisticated jump-diffusion processes that are useful for addressing a wide range of applications, including financial economics (e.g., option pricing models), particle physics (e.g., radiation transport), and machine learning (e.g., diffusion maps).

## Spiking Neuromorphic Hardware Shows Neuromorphic Advantage on Simulating Random Walks

Random walk solutions are often an attractive option for large scale modeling efforts since independent RWs can readily be computed in parallel. Countering these benefits is the large number of RWs required to approximate solutions via a Monte Carlo method, and translating large-scale RW-based particle codes to GPU-heavy computing platforms is an active area of research [16, 17]. Our approach leverages two key features of spiking neuromorphic hardware – the parallel computation of neurons and the event-driven spiking communication between them – to perform a highly efficient mapping of stochastic processes. While deterministic numerical solutions of PIDEs often rely on relatively few large complex calculations, RWs typically rely on many simple computations. As we show, these computations can be efficiently implemented within circuits of spiking neurons.

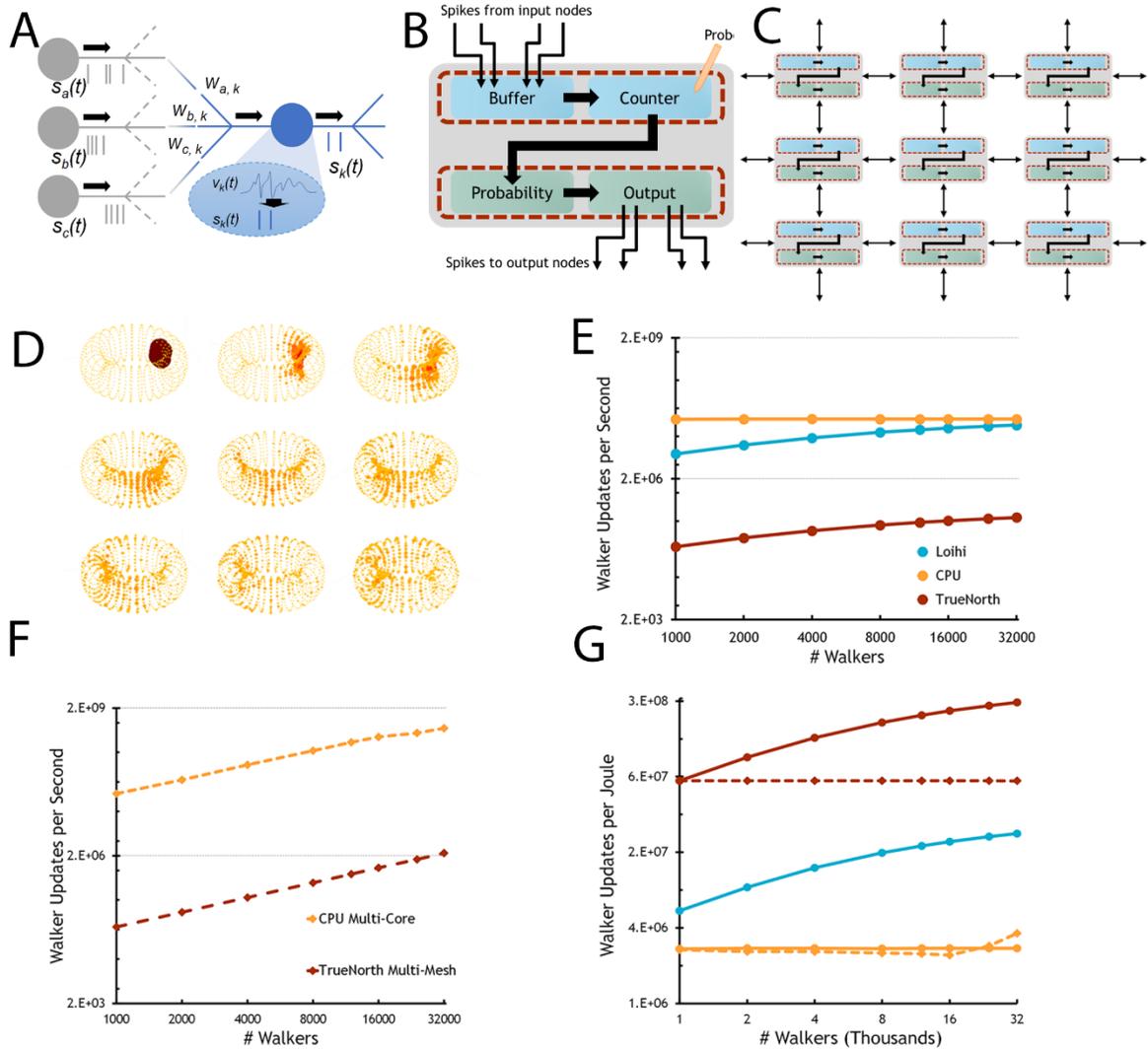

*Figure 1: Neuromorphic hardware can efficiently implement random walks. (A) Leaky integrate-and-fire (LIF) neurons on spiking neuromorphic hardware integrate activity from many inputs, generate a 'spike' if an internal threshold is crossed, and only communicate to targets if the spike exists. (B) Random walk transitions can be performed and tracked by a counter circuit combined with a stochastic output. Each circuit typically comprises of between 10 and 20 LIF neurons, depending on the number of edges. (C) Random walk transition circuits are repeated for every mesh point, and the graph of mesh points equates to the state transition matrix of a discrete time Markov chain. The NMC algorithm implements both the stochastic and deterministic state transitions of all random walkers at all mesh points in parallel. (D) Demonstration of simple diffusion on a 30x30 torus on the Intel Loihi platform. Aside from reading out*

*intermediate states for visualization, the entire random walk process was performed within the NMC system. (E) Simulating additional walkers on Intel Loihi and IBM TrueNorth increases time efficiency, whereas additional walkers have same cost on single CPU core. (F) Random walk processes can be distributed over multiple meshes on IBM TrueNorth or multiple cores on CPUs. (G) The NMC platforms, TrueNorth and Loihi, have a considerably higher energy-efficiency (walker updates per Joule) than CPUs Legend the same as panels E and F. All scaling experiments had 10 replicates with standard errors below 0.5%, so error bars are not shown.*

Our neural algorithm for RWs is based on a previously described circuit to model diffusion, in which we demonstrated that neural circuits can simulate random walks, but did not show how this process could extend to efficiently solve PDEs [34]. In this algorithm, each mesh-point consists of a simple neural circuit that uses common leaky-integrate and fire (LIF) neurons (**Fig. 1A**) to count the number of incoming spikes and a circuit to stochastically distribute spikes to output nodes (**Fig. 1B**). These nodes are then assembled into a graph whose edges represent the transition probabilities from one state to another (**Fig. 1C**). An initial count of walkers is set at the appropriate starting location mesh point (either through input spikes or an initial voltage condition), and once the supervisor circuit initiates the model, the spikes' propagation through this mesh directly reflects the movement of RWs through the corresponding state space. Stated differently, the NMC hardware implements both the stochastic and deterministic components of the stochastic process.

Importantly, this neural algorithm can be generally configured to represent any time-homogeneous Discrete-Time Markov Chain (DTMC) by configuring the shape of this graph and setting the output probabilities within each node to represent the problem description. For instance, a nearest-neighbor mesh with uniform probabilities would lead to Brownian motion in the limit as the mesh and time step go to zero (**Fig. 1D**). More sophisticated RWs, including those with non-local and jump diffusion, walker absorption and creation, can readily be implemented with location-dependent transition probabilities in this framework, allowing the algorithm to realize the processing-in-memory advantages of NMC.

We first performed scaling studies to assess the computational costs inherent in simulating RWs on two NMC platforms, IBM TrueNorth and Intel Loihi, relative to a commodity server-class Intel Xeon E5-2662 CPU. The benchmark task, uniform diffusion on a small torus, was selected to be a simple "best-case" for conventional platforms; we expect any added complexity, such as non-uniform transition probabilities and larger mesh, to preferentially benefit the NMC implementations. For single-threaded implementations of the benchmark task, the CPU is faster than both Loihi and TrueNorth, however Increasing the density of RWs on the NMC platforms required relatively less additional time, whereas additional walkers scale linearly on the CPU (**Fig. 1E**). Distributing the RWs over multiple threads showed comparable time-scaling on multiple cores on CPUs and replicated meshes on TrueNorth (**Fig. 1F**). Despite the slower base clock-rates of these less technologically mature NMC platforms, TrueNorth and Loihi exhibit better single-mesh scaling than conventional platforms and similar multi-threaded capabilities compared to conventional platforms (**Fig. 1G**). Combined, these scaling results satisfy our weaker condition for a neuromorphic advantage.

We next compared the total energy cost of the RW calculations on NMC and conventional platforms for equivalent amounts of computational work. To estimate energy, we scaled the top power estimates for each platform by the relative percentage of the chip used (e.g., number of cores or threads) and integrated over the total simulation time. TrueNorth and Loihi implementations show both a considerable absolute advantage and preferential scaling in total walker-updates-per-Joule compared to the CPU (**Fig. 1g**), satisfying our strong condition for a neuromorphic advantage. Notably, Loihi and TrueNorth appear to occupy different places on the energy-time trade space, possibly in part due to Loihi's incorporation of conventional processors on chip.

Neuromorphic-compatible random walks apply to broad class of PIDEs

While RW solutions to PIDEs have mixed appeal to conventional computing programmers, they have been utilized to provide solutions in a variety of fields, including computer science, physics, medicine, and operations research [23].* The decision between using a deterministic approach and a random walk approach is a complicated and important question. However, this question is beyond the scope of this paper. Rather, we aim to demonstrate that NMC can efficiently implement random walks and, consequently, are able to solve a variety of PIDEs, while potentially mitigating some of the disadvantages of RW solutions (such as the high costs associated with the required number of walkers).

The connection between RWs and the heat equation is well-known. Einstein's 1905 work posits that there exist particles small enough that they may be viewed (with a microscope) but large enough that their Brownian motion is measurable, further arguing that such particles exert a measurable thermodynamic force [10]. Langevin related the mean squared displacement of Einstein's particles to a differential equation describing the particle's motion [21]. A more detailed discussion of the history of this fundamental relationship can be found in [14].

To motivate the probabilistic solution for a larger class of PIDEs, we explore the heat equation. Consider the one-dimensional heat equation with initial condition given by $f(x)$:

---

* Due to their broad relevance, terminology such as "Monte Carlo", "random walks", and other terms may have specific meanings in some fields, so to give clarity to the methods that follow, we emphasize that we are employing discrete-time, finite state space Markov chain approximations to stochastic processes underlying particular PIDEs. These Markov chains are used to generate several random walks. These random walks are evaluated in a Monte Carlo fashion to estimate an expectation.

$$\frac{\partial}{\partial t}u = \frac{1}{2}\frac{\partial^2}{\partial x^2}u, x \in \mathbb{R}$$

$$u(0,x) = f(x).$$

*Equation 1*

Let $W(t)$ be a standard Brownian motion on $\mathbb{R}$. The key relationship relates an expectation (i.e. expected or average value) involving $W(t)$ with the solution $u$:

$$\mathbb{E}[f(W(t))|W(0) = x] = \frac{1}{\sqrt{2\pi t}}\int f(y)\exp\left(-\frac{(y-x)^2}{2t}\right)dy = u(t,x).$$

*Equation 2*

In words, the expectation of a function evaluated at Brownian motion is exactly the solution to the one-dimensional heat equation. This probabilistic representation allows us to approximate the function $u(t,x)$ using RWs. Traditionally, this is accomplished by employing some sampling procedure to generate sample paths of $W(t)$, typically involving a discretization of time and value sampling over a continuous space [15]. Discussed in detail later, in order to make this process amenable to our neural RW algorithm, we must sample our paths through a DTMC $X(j\Delta t)$ that approximates the process $W(t)$. For each spatial location $x_i$, several RWs starting at $x_i$ are generated from the Markov chain. Letting $X_{m,i}$ represent the $m^{\text{th}}$ RW generated starting from location $x_i$, the Monte Carlo approximation gives

$$u(j\Delta t, x_i) = \mathbb{E}[f(W(j\Delta t))|W(0) = x_i] \approx \frac{1}{M}\sum_{m=1}^{M} f\left(X_{m,i}(j\Delta t)\right).$$

*Equation 3*

Regardless of modifications needed for NMC implementation, this simple result can be extended to a more computationally challenging set of problems. Consider the family of PIDEs defined by the equation

$$\frac{\partial}{\partial t}u(t,x) = \frac{1}{2}\sum_{i,j}(aa^\top)_{i,j}(t,x)\frac{\partial^2}{\partial x_i \partial x_j}u(t,x) + \sum_i b_i(t,x)\frac{\partial}{\partial x_i}u(t,x)$$

$$+\lambda(t,x)\int_Q \bigl(u(t,x+h(t,x,q)) - u(t,x)\bigr)\phi_Q(q;t,x)\mathrm{d}q$$

$$+c(t,x)u(t,x) + f(t,x), x \in \mathbb{R}^d, t \in [0,\infty).$$

*Equation 4*

As with Eq. 1, there is an underlying stochastic process, albeit slightly more complicated than just Brownian motion. The stochastic process related to this PIDE is

$$\mathrm{d}X(t) = b\bigl(t,X(t)\bigr)\mathrm{d}t + a\bigl(t,X(t)\bigr)\mathrm{d}W(t) + h(t,X(t),q)\mathrm{d}P\bigl(t;Q,X(t)\bigr).$$

*Equation 5*

The process $X(t)$ is defined by a drift, diffusion, and a non-local jump. In this form, $b$ gives the drift and $a$ gives the diffusion. The process $W(t)$ is a Brownian motion with respect to the underlying space, in this case $\mathbb{R}^d$. The term $P(t;Q,X(t))$ is a Poisson process with parameter given by $-\int_0^t \lambda(s,X(s))\mathrm{d}s$ and the function $h$ describes the non-local jump awarded whenever the Poisson process fires. This stochastic process is readily visualized in **Figs. 2a-c** for constant values of $b$, $a$, and $h$. The jump value $h$ need not be constant and can even be random as seen in **Fig. 2d** ($Q$ can be interpreted as a random variable corresponding to the random jump mark amplitude of a compound Poisson process). The final two panels showcase when the jump value is drawn uniformly over $\{-3,-2,\ldots,2,3\}$. We note that while $c$ does not appear in Eq. 5, it can often be interpreted as an absorption or killing term, demonstrated in **Fig 2e**. A discussion on this interpretation can be found in **SN2**.

Pairing Eq. 4 with the initial condition $u(0,x) = g(x)$, under certain conditions the solution to the initial value problem can be represented as

$$u(t,x) = \mathbb{E}\left[g(X(t))\exp\left(\int_0^t c(s,X(s))\,ds\right) + \int_0^t f(s,X(s))\exp\left(\int_0^s c(\ell,X(\ell))\,d\ell\right)ds \,\bigg|\, X(0) = x\right].$$

<p align="right"><em>Equation 6</em></p>

A proof for the one-dimensional case can be found in **SN2**.

Various special cases of this result exist. A particular interesting special case arises when considering the steady-state version of Eq. 4, where $\frac{\partial}{\partial t}u = 0$ and $t$ does not appear as an argument in all functions. Setting $c = 0$ and considering this case as a boundary-value problem with $u(x) = v(x)$ on the boundary of some domain $D$, the solution can be shown to take the form

$$u(x) = \mathbb{E}\left[v(X(T_x)) + \int_0^{T_x} f(X(s))\,ds \,\bigg|\, X(0) = x\right].$$

<p align="right"><em>Equation 7</em></p>

Here, $X(t)$ is the process given by Eq. 5 with $t$ omitted as the first argument in $a$, $b$, and $h$. Since time is still an argument for the process, the probabilistic solution requires the use of the stopping time $T_x$, or the time for which the random process $X(t)$, starting at $X(0) = x$ exits the domain $D$. A proof for the one-dimensional case can be found in **SN2**.

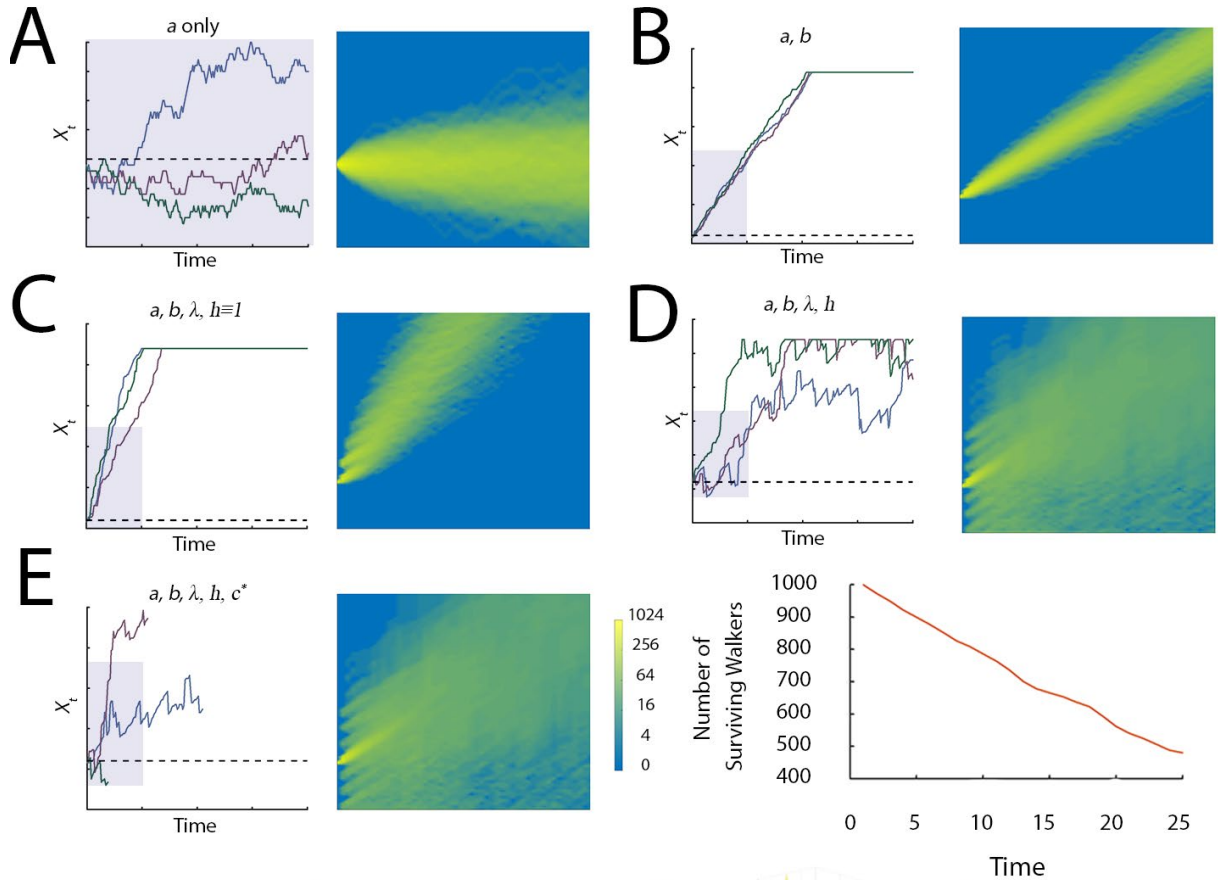

*Figure 2: Random walk processes are well-suited for NMC, and the inclusion of different terms in the stochastic process yields random walks with differing behavior. For (A)-(E), left panel shows three illustrative random walks for 2 seconds (100 time steps); right panel shows density of 1000 random walkers run on Loihi. The range shown in the density plots is highlighted in overlaid onto the process examples. (A) Including only an $a$ term yields basic diffusion; (B) Including $a$ and $b$ yields diffusion with drift. (C-D) The inclusion of $\lambda$ and $h$ allows the random walk to 'jump' for discontinuous movements. (E) The $c$ term under some conditions can yield walker removal. In all plots, the finite range is applied by imposing an upper and lower bound for the walks. (F) Sources of discretization in all stochastic processes (of either conventional or neuromorphic sources) impacts the accuracy and convergence of expectation solution for the PIDE. The first row details the Monte Carlo order of convergence; the second row is the order of convergence for the Euler-Maruyama discretization method; the third row is a best-case scenario estimate for error accrued due to discretizing space; the fourth and final rows merely indicate that some problems could have additional error due to enforcing a finite state space or due to reduced precision on neuromorphic platforms. For further discussion, see* **Methods**.

| **Non-Zero Terms in Eq. 4** | **Example Application** |
|---|---|
| *Time-dependent problems* | |
| $a, b, c, f$ | Stock Option Pricing **[5]** |
| $\lambda, b, c, f, h$ | Boltzmann Flux Density **SN3** Reduced Problem, **Fig. 3A-D**. |
| $a, c$ | Heat Equation with Dissipation (See **Fig. 4C**) |
| *Steady-state problems* | |
| $a, f$ | Electrostatic Scalar Potential, Heat Transport, or Simple Beam Bending [38] |
| $\lambda, b, c, f, h$ | Particle Fluence **SN3** Reduced problem, **Fig. 3E-I**. |

*Table 1: Examples of applications involving a PIDE in the form of Eq. 4. This table is not exaustive and includes only a sample of possible applications. In this paper, we utilize a random walk method to solve two heat transport*

*problems and a reduced problem for both the Boltzmann particle angular flux density problem and the angular fluence problem.*

These PIDEs are important within many application domains, including particle physics, quantitative finance, and molecular dynamics, among others. When viewed probabilistically, the steady-state problems are particularly interesting for neuromorphic because the long run-times required for RWs to reach steady-state solutions are often computationally prohibitive on conventional hardware.

The preceding discussion on the two families of PIDEs and their probabilistic solution representations are largely known results – we merely reformulate these results in forward time (see **SN2**). The new contribution we provide is the use of well-understood DTMC approximations to SDEs in order to make the probabilistic sampling of paths viable on the NMC diffusion algorithm.

A DTMC approximating Eq. 5 is compatible with the neural algorithm we described for diffusion (**Fig 1D**). In particular, the drift $b$ and non-local diffusion terms $\lambda$ and $h$ can naturally be reflected within the definition of the mesh and transition probabilities (**Fig 1C**), in effect providing those extensions to diffusion. Similarly, non-conservation of walkers (walker absorption or creation) can be easily integrated into the system we described. Such a situation may be desirable when the form of $c$ lends itself towards an absorption interpretation.

To approximate Eq. 5 with a DTMC, one must employ some sort of temporal and spatial discretization scheme. Having NMC approximate the DTMC introduces additional sources of uncertainty (**Fig 2F**). Specifically, the finite node structure of NMC architectures forces the DTMC to have a finite state space. In one dimension, this equates to having a maximum and minimum value in the state space. The error of enforcing a finite state space for the DTMC would vary from application to application. The discrete state space arising from the DTMC also introduces error depending on the problem at hand. If the state space of the random walk is already discrete, it introduces no error. In the continuous case, it could introduce error on the order of $\frac{1}{2}\Delta t \Delta s$ on each time step in a special best-case scenario (see **Methods**). Additional error could arise from hardware specific limitations. For instance, the IBM's TrueNorth and Intel's Loihi pseudo-random number generators that we use are effectively limited to 8 bits.

Both conventional simulations, which model each random walker independently and track the evolution of state variables, and our neuromorphic simulations, which model the parallel evolution of random walkers over a state-space represented by the neural circuit, are impacted by each of these error sources. However, the high numerical precision of conventional processing minimizes the impacts of discretizing the values and ranges of state variables, making the dominant errors due to time discretization and the number of random walkers. In contrast, our neuromorphic implementation enables a very large number of walkers at negligible cost, but the dedication of neurons to explicitly representing state variables raises the cost of reducing the meshing error. The implication of these errors will differ considerably across applications in practice.

Results/Examples

To demonstrate the ability of neuromorphic hardware to implement the DTMCs required for solving these PIDEs, we provide a handful of examples. These are grouped into two main categories: particle equations and geometries. The results of our simulations on hardware and spiking neuron simulators can be found in **Fig. 3** and **Fig. 4**. We cover the more salient points of these examples in the next two subsections and relegate the remaining details to **SN3**.

Neuromorphic hardware can simulate particle transport

First, we showcase two examples of particle transport equations with probabilistic representations suitable for our spiking algorithm. The first is an initial-value time-dependent problem detailing the angular flux density of a hypothetical particle (**Fig. 3A**). Consider a hypothetical particle that has a property called 'direction'. This direction property takes on the value $+1$ or $-1$. According to a Poisson process with rate $\sigma_s$, the particle can experience a 'scattering' event. When a scattering event occurs, the particle chooses a new direction with uniform probability. A second Poisson process with rate $\sigma_a$ controls when the particle is absorbed and ceases to exist. These rates correspond to $\lambda$ and $c$, respectively, in Eq. 4. The function $h$ is represented by the change in direction the particle experiences after a scattering event. Coupled with an initial condition $g$, a population of these particles is assumed to obey the Boltzmann equation for angular flux density (see **SN3**).

The angular flux density, $\Phi(t, \Omega)$, is a function of both time $t$ and direction $\Omega$. We will leave the PIDE in **SN3**, but it takes the form of **Eq. 3** with $a, b$, and $f$ all equal to zero. Assigning some initial condition $g$, the solution is given by

$$\Phi(t, \Omega) = \mathbb{E}\left[e^{-\sigma_a t} g(Y(t)) \big| Y(0) = \Omega\right],$$

$$dY(t) = \omega_{Y(t)} dP(t).$$

*Equation 8*

The SDE almost behaves like our hypothetical particle. The 'direction' at time $t$ is given by $Y(t)$. $P(t)$ is a Poisson process with parameter $\sigma_s t$, and $\omega_{Y(t)}$ is the random change in direction of the random walk after a scattering event given the previous direction. The direction remains the same until the Poisson process fires (signaling a scattering event). Once this occurs, the value of $Y(t)$ increments by the random change in direction $\omega_{Y(t)}$. Notably, the random process differs from the hypothetical particle in that it does not account for absorption. Instead, absorption is resolved through the exponential term in the expectation.

We deployed a neural circuit of a DTMC approximating the dynamics of the stochastic process $Y(t)$ on TrueNorth. The description of the random walk and the parameter values used can be found in **SN3**. In this scenario, an analytic solution exists. **Fig. 3C** shows that the true solution is well approximated by sampling just 1000 random walks per each starting condition. Moving to 10,000 RWs per starting position (**Fig 3D**), we see notable improvement in approximation.

This simplified example of particle transport has broad implications. Directly, if we can well-approximate the analytic solution for this reduced particle transport problem, then it will be possible to approximate more complicated particle transport problems where no solution is available. To that end, we have examined a second particle transport inspired example for which no analytical solution is readily available.

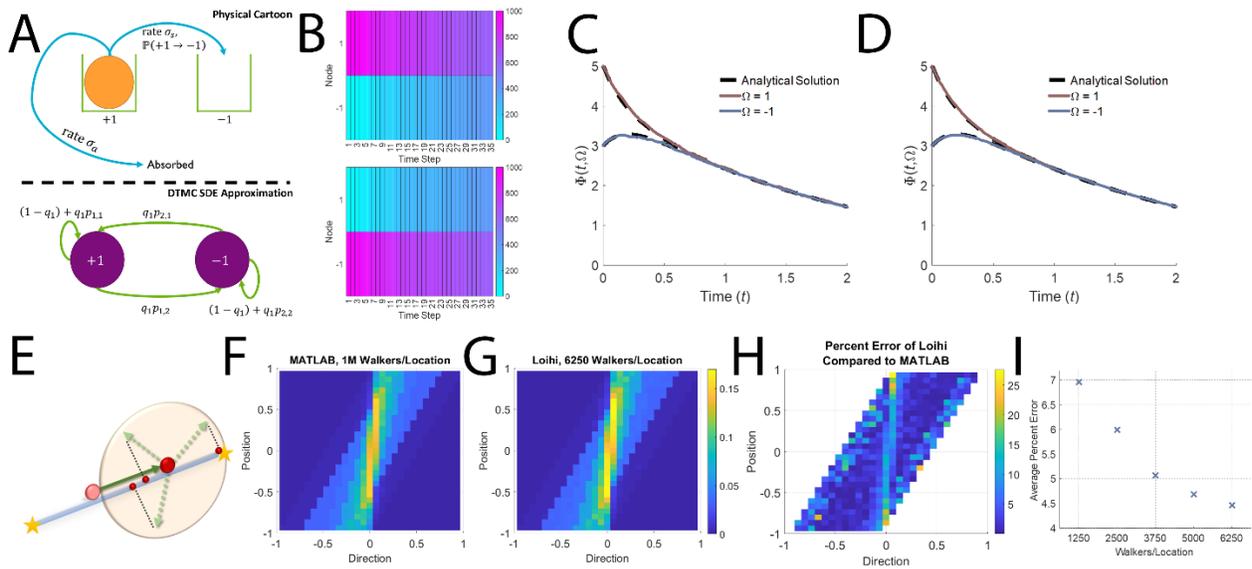

Figure 3: Monte Carlo particle transport simulations on neuromorphic hardware. (A) Non-spatial Boltzmann transition/absorption model (top). Corresponding DTMC approximation for underlying SDE (bottom). (B) Evolution of particles through Boltzmann transitions on TrueNorth. Pink represents higher density of walkers and blue represents lower density for the case where 1000 walkers start in +1 state (top) or -1 state (bottom) and equilibrate due to Boltzmann transitions. (C) PIDE solution calculated through TrueNorth spike data starting 1000 random walkers on each direction. (D) PIDE solution calculated through TrueNorth spike data starting 10000 random walkers on each direction. (E) Spatial particle transport model. Particles travel at fixed speed in measured dimension. At position '0' (red dot), the particles scatter at a random angle preserving their total velocity. At the next time step, the particles will have a different position and direction of movement. (F) MATLAB approximate solution from DTMC implementation of spatial particle model, 1 million walkers at each starting location (G) Intel Loihi approximate solution from DTMC implementation of spatial particle model, 6250 walkers per starting location (H) absolute error

*between Loihi and numerical simulation, (I) average percent error between Loihi and numerical simulation as a function of increasing random walkers per starting location.*

In our second example, we consider a similar particle. This hypothetical particle is subject to scattering events according to a Poisson process with rate $\sigma_s$, however the direction can assume any value in $[-1, 1]$ with a uniform distribution. We assume that this particle is not subject to absorption. In addition to direction, this hypothetical particle also has a spatial coordinate. The particle travels at a speed $v$ in the direction $\Omega$ updating its position (**Fig 3E**). We seek to find the angular fluence $\Psi$, or time-integrated flux, in the spatial domain $[-1, 1]$ subject to the source term $S(x, \Omega)$.

In terms of Eq. 4 (and detailed in **SN3**), $\lambda = v\sigma_s$, $f = vS(x, \Omega)$, $b = v\Omega$, and $h$ is the change in direction after a scattering event given the current direction of travel. The remaining terms, $a$ and $c$, are zero for this example.

Enforcing absorbing conditions on the boundaries, the angular fluence $\Psi(x, \Omega)$ is a function of position $x$ and direction $\Omega$, and obeys the PIDE in **SN3**. The solution may be represented as

$$\Psi(x,\Omega) = \mathbb{E}\left[\int_0^T vS(X(u),Y(u))du \,\bigg|\, X(0) = x, Y(0) = \Omega\right],$$

$$dX(t) = -vY(t)dt,$$

$$dY(t) = \omega_{Y(t)}dP(t),$$

$$T_x = \inf\{t > 0 | X(t) \notin [-1,1], X(0) = x\}.$$

*Equation 9*

Both $P(t)$ and $\omega_{Y(t)}$ are the same as in the previous example. The SDE in this case describes a process with a position given by $X(t)$ and direction given by $Y(t)$. The position updates with velocity $-vY(t)$. The direction only changes by $\omega_{Y(t)}$ whenever the Poisson process $P(t)$ fires.

We deployed a RW approximation from a DTMC of this joint process on Intel's Loihi platform. Details on the DTMC and parameters used are in **SN3**. We completed a 1M walker/location simulation in MATLAB to use as a baseline comparison. One interpretation of the angular fluence is the cumulative density of particles traveling from the source location. From the MATLAB simulation, we see that these particles appear to have mostly traveled with speed $v$ in their original direction assigned by the source, with lessening bands of deviations due to scattering events (**Fig 3F**). Similar to the Boltzmann example on TrueNorth, implementing this simulation on Loihi was able to replicate the numerical examples (**Fig. 3G**) with a low overall error (**Fig. 3H-I**). This low error in the neuromorphic implementation is of particular importance since the low output probabilities due to the high-fan out in this model (up to 30 output nodes) are potentially at risk due to the relatively low 8-bit precision of Loihi's random number generator.

Neuromorphic approach to simulating on non-Euclidean geometries

The particle examples above are straightforward demonstrations of RWs with non-local jumps on a simple domain. We next demonstrated neuromorphic RWs over non-Euclidean domains, solving two heat equations. By carefully defining a mesh and calculating transition probabilities, PIDEs over large complex geometries are no problem for the neuromorphic RW method. To demonstrate the ability of this method to solve problems on non-Euclidean domains, we present two examples involving spheres. While the non-Euclidean shapes we consider are by no means 'complex,' we merely showcase that the method is mostly agnostic to the domain.

Consider a basic heat equation on the unit sphere. We let $\mathbb{S}^2$ represent the unit sphere and take $\boldsymbol{a}(t,\boldsymbol{x})$ to be the positive scalar $\alpha$ for all $t \in [0,\infty)$ and $\boldsymbol{x} \in \mathbb{S}^2$. Set the remaining coefficients in Eq. 4 to zero. Paired with an initial condition $g(\boldsymbol{x})$, the probabilistic solution to the heat equation on the sphere is given by

$$u(t,\boldsymbol{x}) = \mathbb{E}\big[g\big(\boldsymbol{X}(t)\big)\big|\boldsymbol{X}(0) = \boldsymbol{x}\big],$$

$$\mathrm{d}X(t) = \sqrt{2\alpha}\mathrm{d}W(t),$$

*Equation 10*

where $W(t)$ represents Brownian motion on the surface of the sphere. By choosing a particular initial condition, this problem has a tractable analytic solution (**SN3**). The initial condition selected resembles a soccer ball pattern (**Fig. 4A**).

To employ our neuromorphic approach, we must be able to approximate Brownian motion on the surface of the sphere with a DTMC. There are several ways to describe Brownian motion on the sphere, including using the von-Mises Fisher distribution [40], employing representations with spherical coordinates [6], or limiting from higher dimensions [8]. Since it is applicable to other curved shapes, we elect to use a tangent plane approximation. Setting $\alpha = 42$, we deploy a RW approximating the process $X(t)$ on Intel's Loihi platform. Starting 3000 RWs on each position yields the approximate solution found in **Fig. 4A** for a collection of time points. We would expect better agreement with a greater number of nodes and walkers per starting location. Further, as shown in **Fig. 4B**, we see that the low precision of probability transition has acutely increased the amount of error accrued for this example when compared to a MATLAB simulation.

This example provides compelling evidence that complex geometries where analytic methods are less tractable represent an opportunity for NMC impact. These could arise where domains are more complicated than just a single sphere. One could imagine an object with many spines or with several crevices. To start down this path, we present an initial-value problem on the surface of a barbell shaped object.

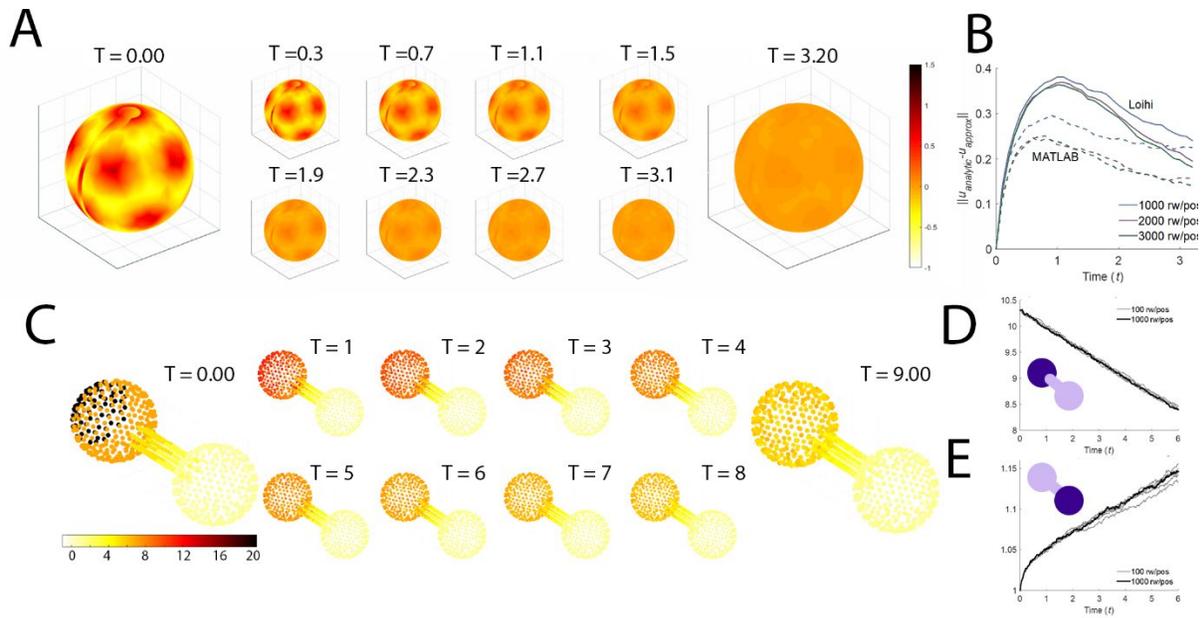

*Figure 4: NMC random walk algorithm can implement random walks over non-Euclidean geometries. (A) Time-course of random walks simulated on Loihi to model heat diffusion on the surface of a sphere. Red locations represent higher initial temperature relative to yellow locations. Heat is conserved on this simulation. (B) Absolute norm of error is higher on NMC relative to MATLAB simulation at initial timepoints, but approaches conventional error levels as simulation progresses. (C) Time-course of random walks run on neural simulator for heat diffusion on two spheres connected by a tube ("barbell"). Heat was allowed to dissipate from the surface. (D) Average temperature of the left sphere decreases rapidly during the simulation. (E) Temperature gradually increases sharply for small time on the right sphere. As time increases, this rate of increase slows as cooling begins to take effect. For large time, the temperature on the right sphere will decrease to zero.*

Consider the heat flow on a barbell with cooling and an initial condition. Let $\mathbb{B}$ represent the surface of the barbell shape. Again, we set $a(t,x) = \alpha$, some positive scalar. Then, to account for cooling, we take

$c(t, \boldsymbol{x}) = \kappa$, another positive scalar. All other coefficients in Eq. 4 are assumed to be zero. Again letting $g(\boldsymbol{x})$ be an initial condition, the probabilistic solution is

$$u(t,\boldsymbol{x}) = \mathbb{E}\big[e^{-\kappa t}g(\boldsymbol{X}(t))\big|\boldsymbol{X}(0) = \boldsymbol{x}\big],$$

$$\mathrm{d}\boldsymbol{X}(t) = \sqrt{2\alpha}\mathrm{d}\boldsymbol{W}(t),$$

*Equation 11*

where $\boldsymbol{W}(t)$ now represents Brownian motion on $\mathbb{B}$. Our discretization of the shape required 748 mesh points (more details on the mesh construction and DTMC are in **SN3**). Due to the mesh-size relative to the currently limited neuromorphic chip sizes available to us, we deployed this example on a spiking net simulator. We implemented a random walk approximating the stochastic process. The results of simulation for various time points can be found in **Fig 4C**. The temperature equilibration of the left (**Fig. 4D**) and right (**Fig 4E**) sides of the barbell proceed as one would expect from thermodynamics.

## Discussion

The results here demonstrate that spiking neuromorphic hardware technology is suitable for implementing a scalable energy-efficient approach to solving an important set of numerical computing problems. Neuromorphic hardware is still immature relative to conventional hardware in terms of both physical scale and clock speed, although it already demonstrates considerable power advantages. Here, we show that our neural RW algorithm scales comparably to a parallel CPU approach, allowing us to observe a significant energy advantage in current neuromorphic platforms today while being positioned to take full advantage of large-scale neuromorphic hardware once realized. We further focus our exploration on demonstrating the broad application impact of our algorithm approach, showing that with simple extensions this approach can apply to a wide range of complex application domains.

Notably, the approach taken here does not leverage all the brain-inspired features present in many emerging neuromorphic hardware technologies. For instance, our approach does not leverage learning;

however, we expect that the neural formulation of stochastic processes may make them more amenable to model calibration against experimental observations, and in situ neuromorphic learning may make this process more efficient. Likewise, we focused our demonstrations on large-scale digital spiking platforms, such as Loihi and TrueNorth, because they exist at the requisite neural scales for our algorithms. There is considerable interest in analog neuromorphic approaches that should similarly be compatible with this approach [19, 30, 32], although we would have to consider the precision implications of analog devices alongside the other approximation considerations (**Fig. 2b**).

One important consideration of this work is that the numerical accuracy of our neural approach is relatable to typical numerical precision considerations in conventional computing. Stated differently, this approach avoids the approximation pitfalls associated with many AI algorithms, wherein the implications of numerical precision and interpretability is still an open question. While understanding and accounting for these approximation errors will be critical for any application, the graph-based approach taken here provides several well-understood design choices to tailor the algorithm and hardware solution appropriately given precision concerns. For instance, in applications with clearly defined state spaces, such as diffusion over a social network, the mesh can directly map to the system, and resources can be dedicated to adding more walkers. Alternatively, in complex geometries or unbounded systems, it may be necessary to commit considerable neuromorphic hardware to a larger mesh.

Whatever the eventual set of capabilities that future neuromorphic platforms have, we expect that neuromorphic hardware will eventually exist primarily in heterogeneous system architectures alongside CPUs, GPUs, and other accelerators [20]. The neuromorphic algorithms for solving PIDEs described here complement AI as an application for brain-inspired hardware, and they strengthen the long-term value proposition for neuromorphic hardware in future computing systems. Further, in contrast to neural network applications, where neuromorphic hardware has struggled to match the speed of GPUs and linear algebra accelerators, our work shows that in the realm of numerical computing, neuromorphic hardware not only can deliver concrete energy advantages today, but is capable of scaling effectively in terms of processing time and overall efficiency.

## Acknowledgments

We thank Steve Plimpton and Andrew Baczewski for reviewing an early version of the manuscript and Adam Moody, Suma Cardwell, and Craig Vineyard for managing access to the TrueNorth and Loihi platforms. The authors acknowledge financial support from Sandia National Laboratories' Laboratory Directed Research and Development Program and the DOE Advanced Simulation and Computing Program. Sandia National Laboratories is a multiprogram laboratory managed and operated by National Technology and Engineering Solutions of Sandia, LLC, a wholly owned subsidiary of Honeywell International, Inc., for the U.S. Department of Energy's National Nuclear Security Administration under contract DE-NA0003525.

This article describes objective technical results and analysis. Any subjective views or opinions that might be expressed do not necessarily represent the views of the U.S. Department of Energy or the U.S. Government.

## Contributions

JDS, BF, RL derived mathematical results, JDS and BF designed particle experiments, JDS, WS, and JBA designed geometry experiments, OP, WS and JBA developed the neuromorphic algorithm and performed theoretical neuromorphic complexity analysis, AJH and JBA performed neuromorphic simulations, JDS, LR, and WS performed software simulations, and all authors wrote the paper.

Data Availability Statement

The computational scaling data generated and analyzed in this study are included in the published article as Extended Data.

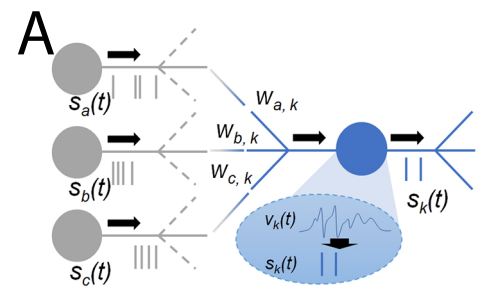 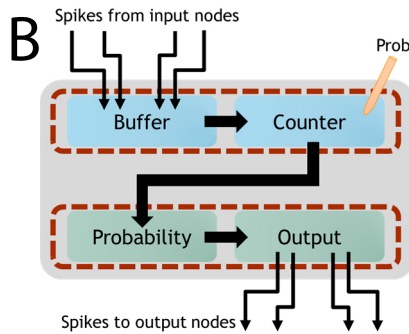 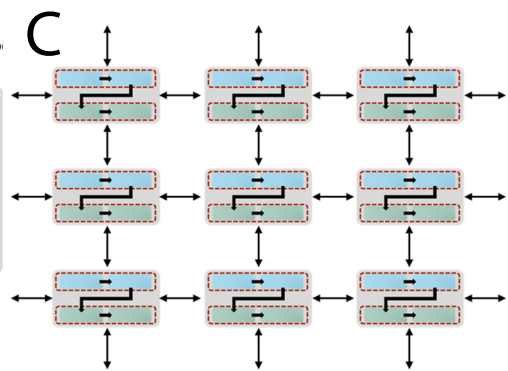
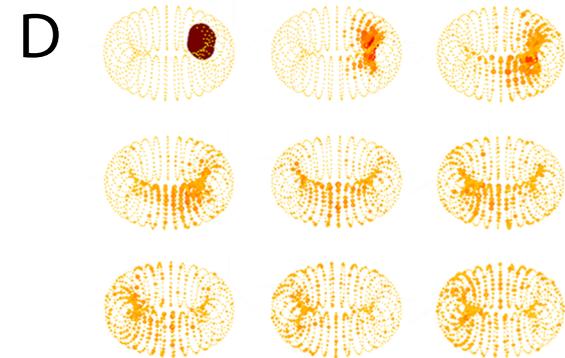 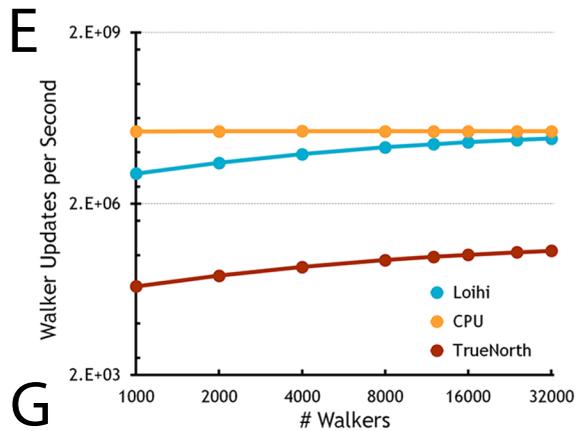
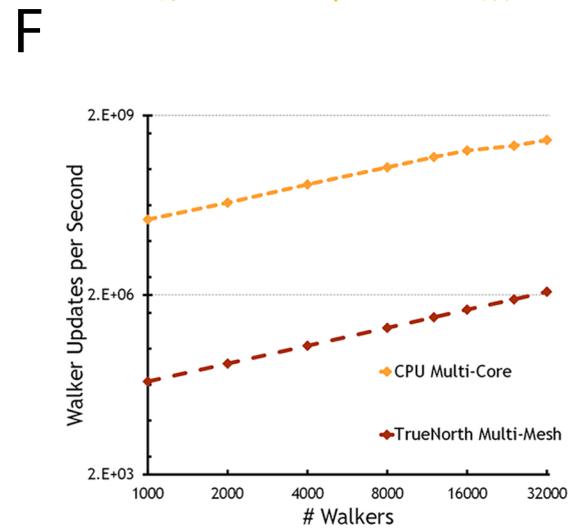 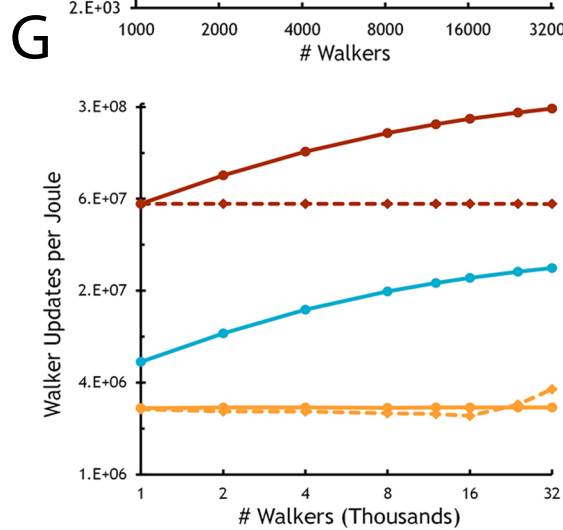

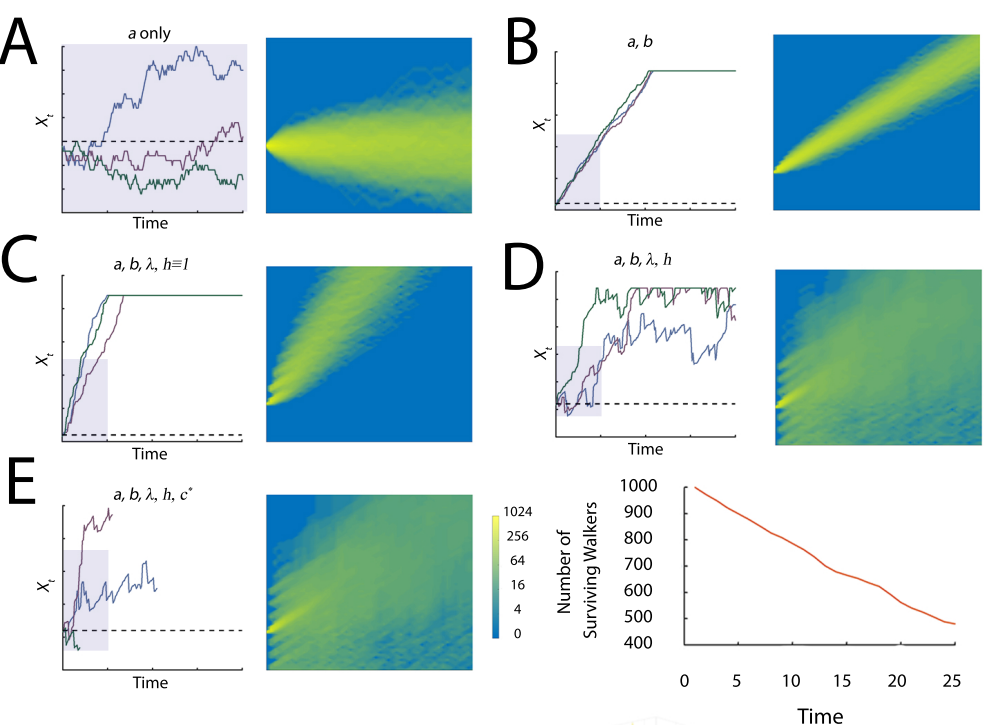

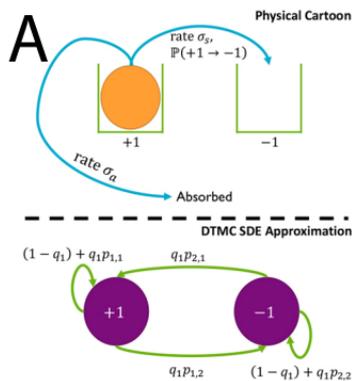 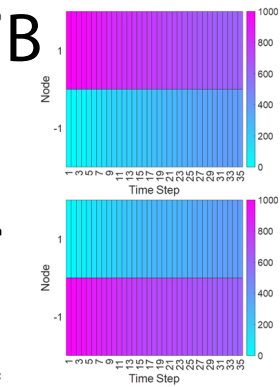 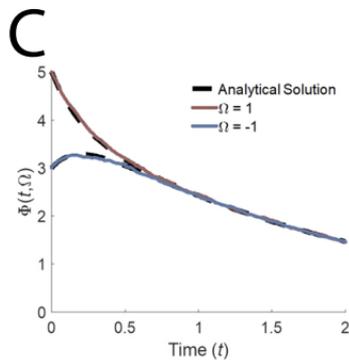 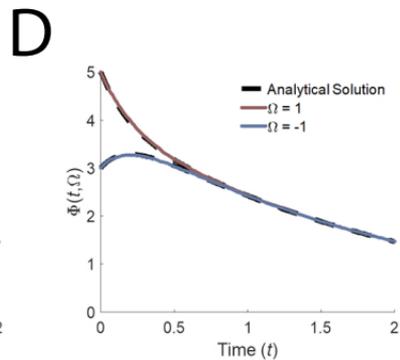
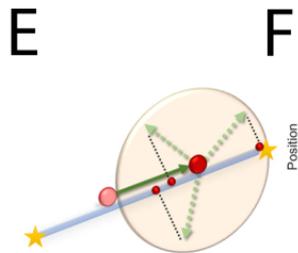 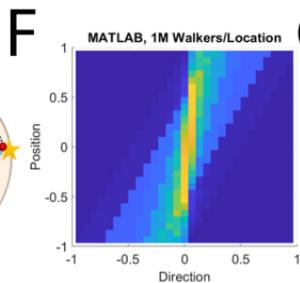 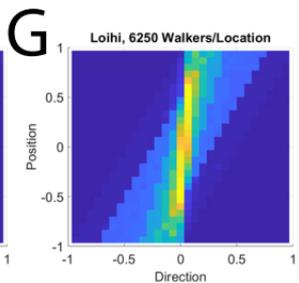 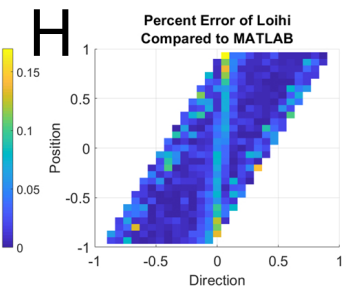 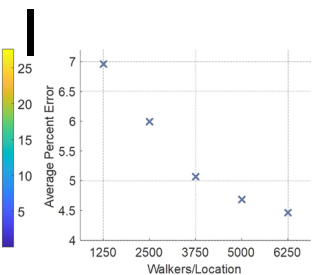

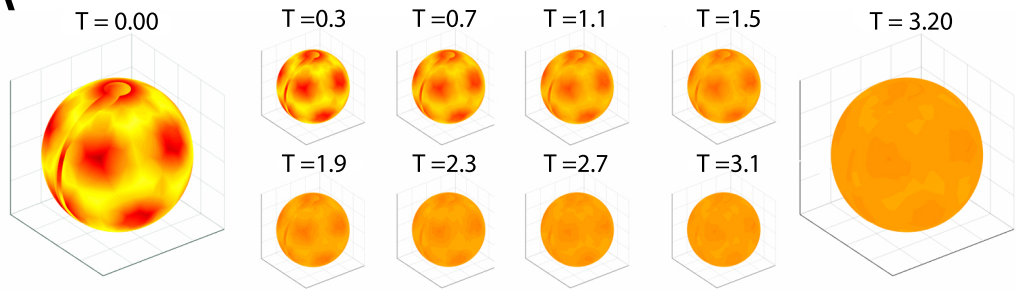
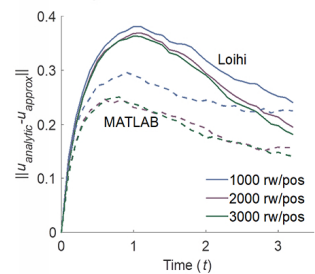
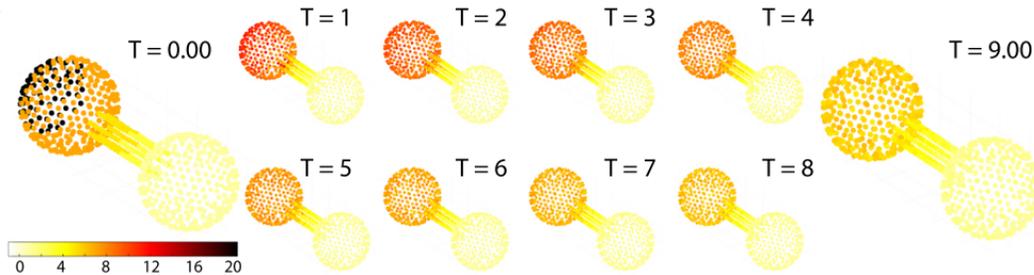
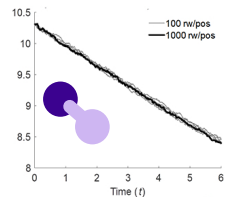
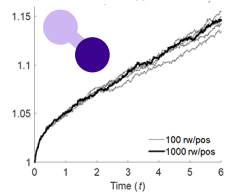

# M-1  Methods

We subdivide our Methods section into two main components: neuromorphic hardware methods and mathematical methods. In Section M.1, we provide details on the random walk circuit as well as model implementation. In Section M.2, we describe one method for approximating a stochastic differential equation with a finite state space, discrete-time Markov chain. We also showcase how to average the random walks to approximate the solution to a PIDE.

## M.1  Neuromorphic Hardware Methods

### General

The neurons used on both Loihi and TrueNorth either are integrate-and-fire neurons (IF) or threshold gate (TG) neurons. In both cases, the hardware neurons integrate all active synaptic inputs and make a decision to fire based on a threshold. For IF neurons, if a neuron does not fire, there is no decay of the neuron's internal voltage. In contrast, TG neurons have full decay every time step, regardless of whether it spikes.

For both TrueNorth and Loihi, the implementation of the random walk algorithm was based on the density circuit described in [14]. Each mesh point in the simulation consisted of two counting circuits (one to buffer inputs, one to count down outputs) and a probabilistic fan-out circuit. The network also utilized a population of supervisor neurons to control the timing and synchrony of the walkers through the circuit. For cases where walkers are synchronized (each time step involves every walker advancing in time by 1), the separate buffer circuit is required to accumulate all walkers coming to a location, however this is unnecessary if the walkers can be run asynchronously. We briefly describe the design of each circuit here in the context of Loihi. The TrueNorth configuration was similar, albeit with some minor differences that are noted.

### Buffer and Counter Circuits

In the random walk algorithm, the buffer circuit and walker counting circuit are identical, with the only difference being the inputs and outputs. The counting circuits on the Loihi are structured as shown in Fig. M.1. Each circuit consists of three neurons: an IF "**count**" neuron, which stores the count of random walkers at that location in its internal voltage (as a negative distance from threshold), a TG "**generator**" neuron, which is designed to spike until the counter neuron reaches its threshold, and a TG "**relay**" neuron, which corrects for situations where there are no walkers at that location. In the buffer circuit, the **count** neuron receives synaptic inputs (weight$= -1$) from other mesh node outputs. In the counter circuit, this neuron receives a synaptic input (weight $= -1$) from the buffer **generator** neuron. In both cases, the **count** neuron will represent the cumulative walker density at that location. The output of the **count** neuron is an inhibitory connection to its respective **generator** neuron, designed to stop its activity

The circuit is designed so that when the supervisor activates the circuit, the **generator** neuron will continue to spike until the **count** neuron reaches a threshold and sends an inhibitory spike to stop it from firing. Thus, at each simulation time step, if there are $k$ walkers at a location, the **generator** neuron will fire $k+1$ times (which is subsequently corrected for). During the first half of the simulation time step, the buffer **generator** neuron transfers the count from the buffer to the counter circuit, and during the second half of the simulation time step, the counter **generator** neuron transfers the count to the probabilistic fan-out circuit (described below) which then distributes the walkers to a different mesh node's buffer circuit.

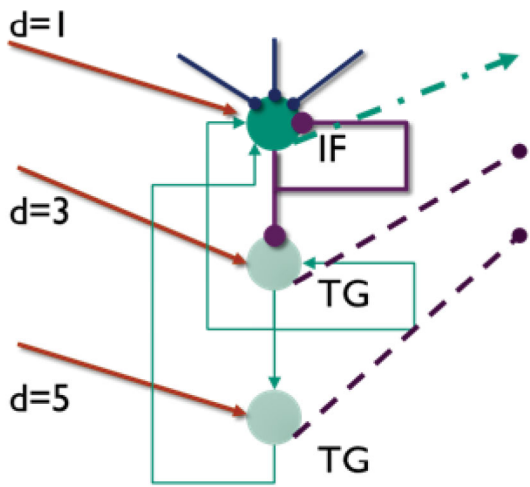

Figure M.1: Neural Circuits for Buffering and Counting on Loihi. Red input lines (from left) represent inputs from supervisor neuron. Circle ends represent inhibitory connections (weight = -1), arrows represent excitatory connections (weight = 1). For buffer circuit, outputs (to right) go to counter circuit **count** neuron; for counter circuit, outputs go to probability neurons.

The **relay** neuron is present to account for the subtle timing between the **generator** and **count** neurons that provides the extra signal from the **generator** neuron as well as help handle cases where the mesh point has no walkers. There are several mechanisms for performing these corrections, which also result in a few additional synaptic connections, for which the Loihi strategy is shown in Fig. M.1.

## Probability Circuit for Loihi

The goal of the probability circuit is to send a walker, as a spike, to one of the mesh node's downstream target's buffer **count** neuron. The circuit is designed to use intrinsic pseudo-random number generators (PRNGs) available to each individual neuron to select **only one** of the mesh node's outputs at the appropriate Markov transition probability.

While there are likely several implementations of a circuit to send a spike to one of $N$ outputs with a probability $p_{\text{out, } i=1,\ldots,N}$, the methods we selected on Loihi and TrueNorth were identified to account for the particular nature of the PRNGs on each chip.

For Loihi, the PRNG provides a random input onto each neuron on a particular neural core as an 8-bit pseudo-random integer, with potential multiplicative and additive scaling. For our purposes

M-58  here, we can consider this random number as an integer uniformly chosen from $[-127, 128]$. Suppose
M-59  we want a neuron, $b$, that has a threshold 100, to fire at probability $p$ when it receives a spike from
M-60  an upstream neuron, $a$. We then can set an input weight from $a$ to $b$ as

$$w_{ab} = 128 \times p_b + 36. \tag{M.1}$$

M-61  To output to one of $N$ neurons, we considered the neural circuit that collapses a probability tree
M-62  into a single layer. For instance, one can consider a uniform 3 layer decision tree, with each branch
M-63  having a 50% probability, leading to eight outputs with 12.5% probability. This scales as a depth
M-64  $\log_2(N)$ circuit with $N - 1$ total probability neurons. We can compress this into a single layer, by
M-65  having each output node (the leaves on the decision tree) requiring that all positive input branches
M-66  are active and receiving inhibition from all "wrong" decision branches, and no inputs from branches
M-67  on the other side of the decision tree.

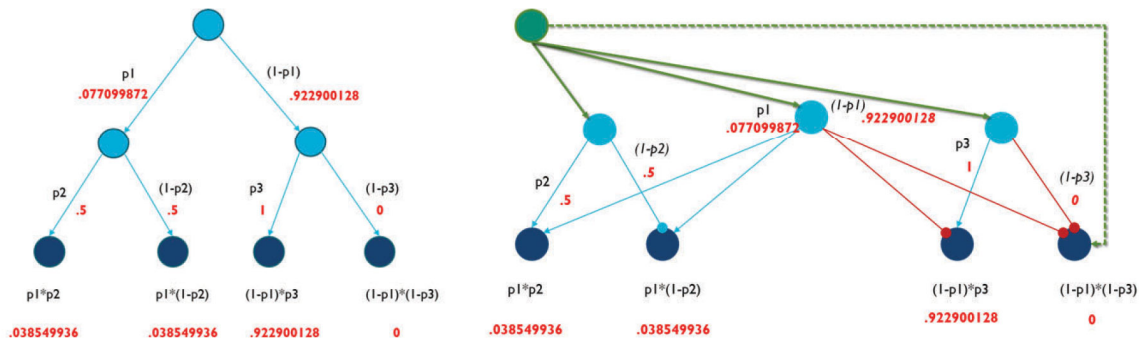

Figure M.2: Illustration of computing probabilistic circuit. Left: hypothetical decision tree to compute probabilities with example output probabilities in red. Right: same decision tree compressed into a single layer, with source input driving probabilistic choice. The dotted line is an excitatory connection with a delay to correspond to skipping the probabilistic layer. From source neuron, weights from source neuron (green) to probability neurons (blue) are set to tune probabilities neurons fire, per equation M.1. Outputs of probability neurons with arrows are excitatory (weight = 1) and with circles are inhibitory (weight = −1).

M-68  Procedurally, this is achieved by having $N-1$ probability neurons, whose probabilities of being
M-69  active on a given time step are given by the following procedure.

---

**Algorithm 1:** Determine Probability Circuit

**Input:** Neuron $g$ that generates a spike per walker
**Input:** Output nodes $o_1 \ldots o_N$ with desired output probabilities $\nu_1 \ldots \nu_N$
// Assume N is a power of 2
$T :=$ A binary tree with $o_1 \ldots o_N$ as the leaf nodes
**for** *non-leaf node* $n = 1 \ldots N - 1$ **do**
  | Label one edge as 'positive' and the other as 'negative'
**end**
**for** *non-leaf node* $n = 1 \ldots N - 1$ **do**
  | $\Lambda := \{i : o_i \text{ is a leaf of } n\}$
  | $\Lambda_p := \{i : o_i \text{ is a leaf of } n \text{ connected through the 'positive' branch}\}$
  | $p_n \leftarrow \sum_{i \in \Lambda_p} \nu_i / \sum_{i \in \Lambda} \nu_j$ // Assign a conditional probability to each node
**end**
// T is a ``probability tree''
$\hat{T} :=$ set of $2N - 1$ neurons
**for** *node $n$ in $T$* **do**
  | $\hat{n} \in \hat{T} :=$ a neuron with threshold 1 and decay 0
  | **if** *n is a leaf node* **then**
  |   | $\mathcal{P} = (n_i, e_i)_{1 \ldots L} :=$ the unique path from the root node to $n$
  |   | $c :=$ number of 'positive' edges in $\mathcal{P}$
  |   | **for** $i = 1 \ldots L$ **do**
  |   |   | **if** $e_i$ *is a 'positive' edge* **then**
  |   |   |   | $weight[\hat{n}_i, \hat{n}] \leftarrow 1/c + \epsilon$ // All 'positive' edges must have spikes
  |   |   | **else**
  |   |   |   | $weight[\hat{n}_i, \hat{n}] \leftarrow -1$ // No 'negative' edges can have spikes
  |   |   | **end**
  |   | **end**
  |   | **if** $c == 0$ **then**
  |   |   | $weight[g, \hat{n}] \leftarrow 1$ // with delay matching depth of circuit
  |   | **end**
  | **else**
  |   | $weight[g, \hat{n}] \leftarrow 1$ // alternatively can be set to control neuron fire probability, as in equation M.1 for Loihi
  |   | $p_{\hat{n}} \leftarrow p_n$ // convert the probability to a probability to fire
  | **end**
**end**

---

M-70

M-71 We then set the threshold to number of positive inputs onto the output neurons. For neu-
M-72 rons that don't have any positive inputs, we include a connection with delay $= 3$ from the spike
M-73 **generator**.
M-74 The output neurons, when activated, will send a single spike to their corresponding mesh node's
M-75 buffer **count** neuron. This corresponds to the transfer of that walker to the different mesh location.

## M-76  Probability Circuit for TrueNorth

M-77    The probability circuits for Loihi and TrueNorth are functionally equivalent, but due to differ-
M-78  ences in the two hardware platforms the realizable TrueNorth circuit differs slightly from that of
M-79  Loihi. It is important to provide adequate coverage of the TrueNorth probability circuit in this
M-80  section for completeness.
M-81    TrueNorth has a few different stochastic neuron dynamics that can be configured for each
M-82  neuron. For this material we are taking advantage of stochastic leak. The two pertinent neuron
M-83  equations that emerge when utilizing TrueNorth's stochastic leak function is the leak equation
M-84  defined as

$$V_j^*(t) = V_j(t) + F\left(\lambda_j, \rho_j^\lambda\right), \text{ where,}$$

$$F\left(\lambda_j, \rho_j^\lambda\right) = \begin{cases} 1 & \text{if } \lambda_j \geq \rho_j^\lambda \\ 0 & \text{otherwise} \end{cases}.$$

M-85  Further, $\rho_j^\lambda$ is a random sample from the PRNG drawn from $\mathcal{U}(0, 255)$ and $\lambda_j$ is an integer in
M-86  the range of $[0, 255]$. The stochastic property of the neuron is ultimately controlled by setting the
M-87  neuron parameter $\lambda_j$. We then set the threshold of this neuron to 1 with a reset potential of 0.
M-88  Thus, if $F$ evaluates to 1 the neuron will fire, and if $F$ evaluates to 0 the neuron will not fire. In
M-89  effect, we created a stochastic neuron that fires with some probability defined by $F$.
M-90    We exploit this stochastic neuron dynamic to provide a neuron circuit that directs a walker
M-91  (spike) through a binary tree to give rise to a controllable exit probability for each of the four exit
M-92  neurons of the mesh node. We define a probability neuron that receives the walker input (spike)
M-93  but also receives input from the stochastic neuron. We set this neuron's threshold at 1 with a leak
M-94  value of $-1$. These probability neurons are depicted in Figure M.3 as $r_0$, $r_1$, and $r_2$. Thus, if a
M-95  walker enters this probability neuron it contributes a value of 1 to the neuron potential and if the
M-96  stochastic neuron also provides a spike, or value of 1, the potential will have a value of 2. Then, the
M-97  probability neuron will leak a value of $-1$ to result in a final potential value of 1 that is compared
M-98  to the threshold value of 1, which will result in a fire of that neuron. This neural dynamic requires
M-99  two spikes to fire, if it only gets 1 spike, meaning the stochastic neuron did not fire, the probability
M-100 neuron will not fire.
M-101   This is the exact formulation that defines the probabilistic nature of the mesh nodes in TrueNorth.
M-102 The leaf nodes of the tree define the exit directions of each mesh node. An easy way to understand
M-103 how the three probability nodes influence the exit direction is to consider the combinatorial effects
M-104 of the probability nodes firing. For $o_0$ to fire it means $r_0$ and $r_1$ fired. For $o_1$ to fire it means $r_0$
M-105 fired and $r_1$ did not. For $o_2$ to fire it means $r_0$ did not fire and $r_2$ did. For $o_3$ to fire it means both
M-106 $r_0$ and $r_2$ did not fire.
M-107   More formally, each leaf node's probability of firing can be related to the probabilities of the
M-108 tree nodes probability to fire. Specifically,

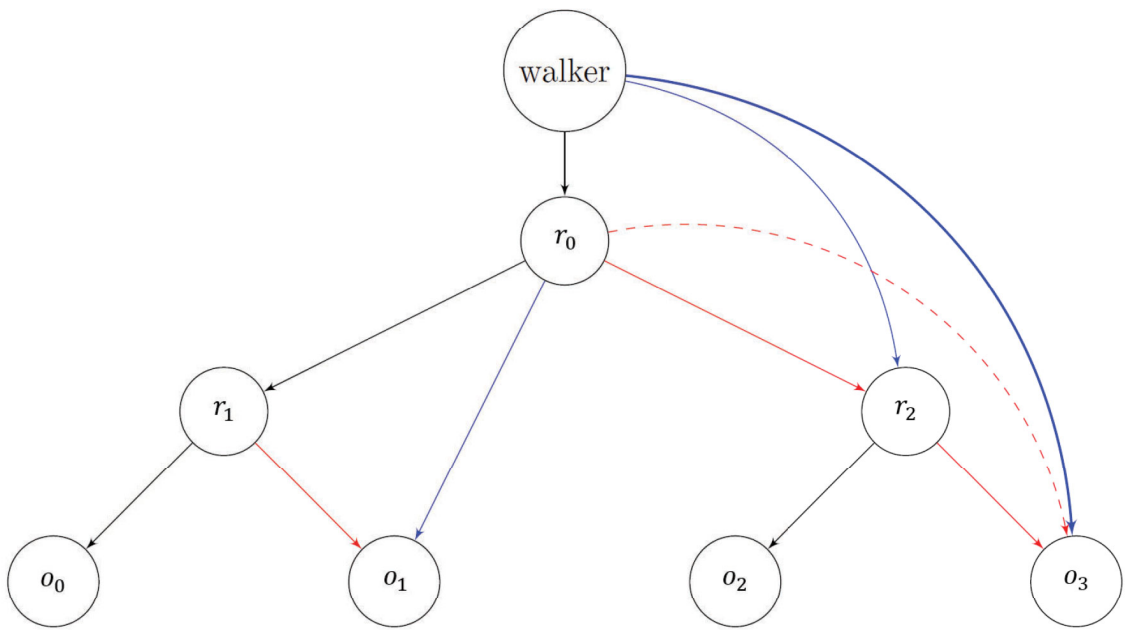

Figure M.3: Binary tree representing the stochastic walk through a TrueNorth mesh node. Probability neurons are $r_0$, $r_1$, and $r_2$. Black edges are excitatory, red edges are inhibitory. Blue edges indicate a delay of 1, and bold blue and red dashed edges indicate a delay of 2. The four leaf nodes, $o_0$, $o_1$, $o_2$, and $o_3$, are the directional nodes with derived exit probabilities.

$$P(o_0) = P(r_0) \times P(r_1),$$
$$P(o_1) = P(r_0) \times (1 - P(r_1)),$$
$$P(o_2) = P(r_2) \times (1 - P(r_0)), \text{ and}$$
$$P(o_3) = 1 - (P(r_0) \times (1 - P(r_2)) + P(r_2).$$

But, in practice we prefer to define the probabilities of each exit direction first and formulate what the stochastic parameter of the stochastic neuron should be. To this end,

$$P(r_0) = P(o_0) + P(o_1),$$
$$P(r_1) = \frac{P(o_0)}{P(o_0) + P(o_1)}, \text{ and}$$
$$P(r_2) = \frac{P(o_3)}{1 - P(o_0) + P(o_1)}.$$

On TrueNorth, this stochastic dynamic utilizes, effectively an 8-bit PRNG. Technically, the PRNG on TrueNorth is 32-bits, but the stochastic leak use case masks off the upper 24 bits and

M-113    delivers the lower 8 bits of the PRNG to the computational circuit of the neuron. From the an-
M-114    alytical side of solving PDEs with random walk algorithms the exit probabilities of each mesh
M-115    node are defined by real numbers in the range of $[0, 1]$. To convert the real valued represen-
M-116    tation to the lower bit resolution, for acceptance into a TrueNorth model, we multiply the real
M-117    value by 256 and then apply a standard round operation to the result. This essentially scales
M-118    the range of $[0, 1]$ to the range of integers from $[0, 256]$. Because the range of $\lambda_j$ is $[0, 255]$, we
M-119    then shift the scaled range left by 1, effectively creating represented probabilities in the range
M-120    of $[-1, 255]$. What is meant by the value of $-1$ is a probability of 0. Since $-1$ is not an
M-121    acceptable value for the TrueNorth model we handle this by removing the stochastic neurons
M-122    synapse to the probability neuron and setting the stochastic neuron's stochastic parameter to
M-123    0. The removal of that synapse effectively provides it a 0 probability of firing. More specifi-
M-124    cally, there is a 0 probability of the stochastic neuron delivering a spike to the probability neuron.

### Implementing Models on Loihi and TrueNorth

M-127    Models are initialized on Loihi and TrueNorth by generating a mesh of equivalent buffer and
M-128    counter circuits and probabilistic circuits tailored to the outputs defined in the application's Markov
M-129    Transition table. Specific to TrueNorth, a connectivity diagram is represented in Figure M.4 which
M-130    provides the implementation details of neuron connectivity for implementing the mesh node in
M-131    TrueNorth; this is the TrueNorth model representation of M.3. Importantly, different mesh points
M-132    can have distinct numbers of outputs, though more outputs will directly equate to more neurons
M-133    required. Further, there is no real restriction on the types of graphs and the connectivity, though
M-134    there will be resource constraints in terms of overall neuron counts and hardware-specific fan-in and
M-135    fan-out constraints, if any. It should be noted that cases where network connectivity restrictions
M-136    are problematic can typically be resolved by replicating target neurons or mesh points.

M-137    For most of the examples in the main text, the Loihi and TrueNorth neuromorphic models are
M-138    unchanged, with the only variable input being the transition matrix used to define the connections
M-139    between mesh nodes and the weights onto the probability neurons in the mesh points. In most
M-140    examples, inputs are given by providing a sequential number of spikes to the appropriate buffer
M-141    **count** neuron of the mesh location from which the random walks will be initialized. There are
M-142    precision considerations in the internal voltage levels that differ between platforms, so we held the
M-143    number of walkers on Loihi to a maximum of about 1000 for the initial condition; although this
M-144    likely can be higher. On TrueNorth, the range of voltage potentials can support up to 393215
M-145    walkers. Full details of generating Markov transition matrices for the examples in the main text
M-146    used on Loihi and TrueNorth are given in **Supplemental Note 3**.

### TrueNorth Scaling Studies: Execution and Statistics

M-148    We performed a number of scaling experiments on IBM's Neurosynaptic System (TrueNorth) to
M-149    better understand how the random walk algorithm performs on actual neuromorphic hardware.
M-150    The base experiment is a random walk simulation on a $21 \times 21$ node torus mesh, where each node
M-151    has 4 incoming and 4 outgoing connections. This corresponds to a walker on the surface moving
M-152    up, down, left, or right. The transition probabilities of the 4 directions are all an equal 0.25. All walkers begin the simulation at the center of the mesh, and the simulations are ran for 100,000 time steps. Here, a time step is the unit of time for all walkers to have moved one position.

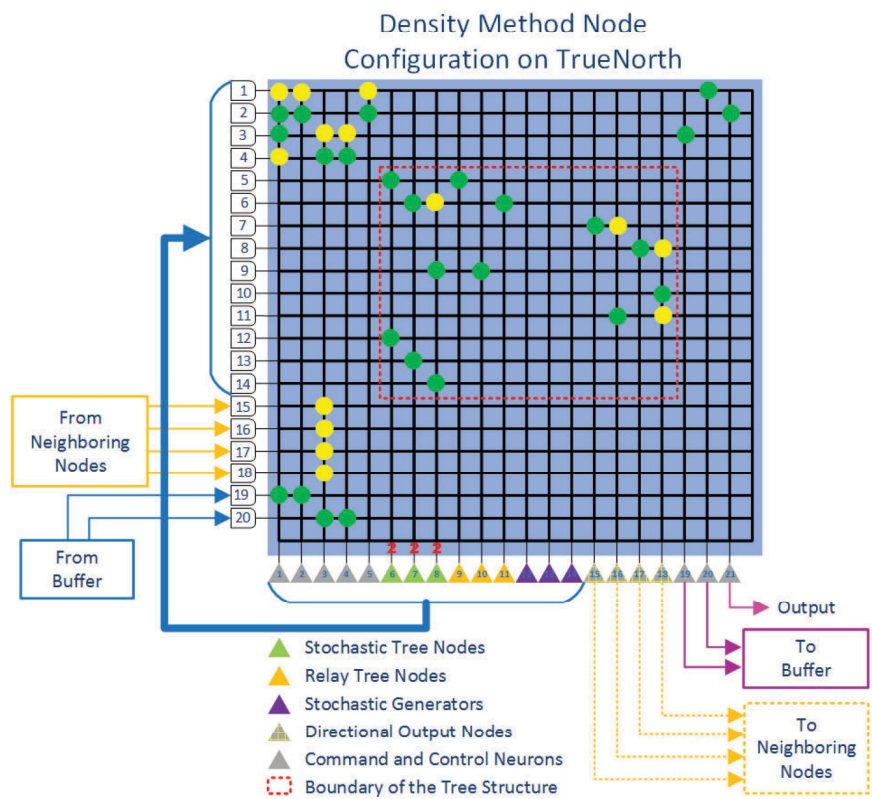

Figure M.4: A near complete specification of the TrueNorth mesh node model for a random walk algorithm. This is a more defined representation of the binary tree from Figure M.3. Neurons are represented by triangles, neuron inputs are on the left edge of the square and a synapse to a neuron is defined by a circle on the cross bar. Green circles are excitatory connections and yellow circles are inhibitory connections. The red number 2 above neurons 6, 7, and 8 indicate that they fire as a result of 2 or more incoming spikes, all other neurons fire as a result of 1 or more incoming spikes.

Initially, the number of walkers is increased from 1000 to 32000 (see **Fig. 1e-g** in the main text, red line) in 8 different runs. Then, this is repeated, increasing the number of walkers through parallelization. Copies are made of the underlying mesh, and each mesh is given 1000 random walkers. The mesh copies increase to keep the total walkers in the simulation consistent with the static mesh runs.

Each experiment had 8 parameter sets that scaled the number of walkers or a parameter of parallelism (mesh copies or number of starting locations). Within each of these parameter sets, 10 trial iterations were executed to allow for a meaningfully computation of the mean and standard deviation of the trial set. For execution on TrueNorth, the total number of neural time steps, or ticks, must be defined *a priori*. Because of the random nature of these experiments, achieving exactly 100,000 time steps is not possible given a fixed number of ticks. Therefore, sample trials were run to achieve a close enough tick count per each parameter set to achieve close to 100,000 time steps. Then the data was normalized to exactly 100,000 time steps by computing the tail time step to tick count ratio. The tail step ratio is defined by taking the mean step ratio of all simulation steps ignoring the first 1,000 simulation steps (this is an arbitrary choice, but, based on empirical evidence, the step ratio flattens out after the first few hundred time steps as the random walk reaches sufficient diffusion). This mean step ratio was then used in conjunction with the tick rate (seconds per tick) to subtract or add time to the measured time of the experiment's trials. For example, if a particular trial ran for 100,034 time steps, had a mean tail ratio of 17.382, an execution time of 937s, and a tick rate of 0.5ms/tick we would take $937 - 34 \times 17.382 \times 0.0005 = 936.705$s as our normalized execution time.

### Computer Simulations

For validation of our algorithm and simulations of the barbell experiment, we performed simple direct simulations of the neural algorithm dynamics. This is a simple direct Python discrete-time simulation of the neural algorithm, not a formal simulator like NEST or a model description language like PyNN. For this, we used the reference simulator described in in "Composing Neural Algorithms in Fugu" [3].

## M.2 Mathematical Methods

Here, we discuss a general approach to creating a discrete-time, finite state space Markov chain approximation to a jump-diffusion SDE.

The success of approximating solutions to PIDEs hinges on the ability to implement a random walk approximation of the underlying stochastic process for each PIDE. While there are many different ways to construct such an approximation, we will discuss the basics and point out where variations can occur.

### M.2.1 Construction

We will explain a basic method for constructing a discrete-time finite state space Markov chain approximation to the following one-dimensional SDE:

$$\mathrm{d}X(t) = b\left(t, X(t)\right) \mathrm{d}t + a\left(t, X(t)\right) \mathrm{d}W(t) + h\left(t, X(t), q\right) \mathrm{d}P\left(t; Q, X(t), t\right). \tag{M.2}$$

Here, $b$ represents the drift of the process $X(t)$ and $a$ represents the diffusion. A non-local diffusion, or jump term, is governed by the process $P$ with reward given by $h$. $Q$ is the jump-amplitude mark random variable with probability density function given by $\phi_Q(q; t, x)$.

For discussion, we assume that $X(t) \in \mathbb{R}$ and that $a$, $b$, and $h$ are such that $X(t)$ could assume any value in $\mathbb{R}$ with non-zero probability. The Markov chain approximation we construct will take values on a finite set. This means we will need to divide the real line into a finite number of intervals to represent our state space and, using (M.2), determine the probability of transition between these intervals.

However, the determination of this probability is subtle – a representative location for the interval is needed, and neuromorphic hardware constraints may limit the number of allowable transitions. Given a starting location, there is a non-zero probability that the process $X(t)$ can transition to any interval on the real line. If the number of allowable transitions in the Markov chain is limited, care must be taken to conserve probability. That is, the particle must transition to somewhere and the probability of the allowable transitions must sum to 1.

We will carefully explore the nuances of the approximation, first exploring a countable state space and then restricting to a finite one. Specifically, we will follow this order:

1. Define a countable state space for the Markov chain;

2. Determine neighbors for each state in the Markov chain;

3. Calculate the probability for each transition in the chain; and

4. Restrict to a finite state space.

The ultimate artifact of construction is a transition matrix among the states of a Markov chain.

### M.2.1.1   Countable state space for the Markov chain.

To begin our approximation for (M.2) we must chop the real line into a sequence of "bins" or "nodes." We note that this discretization does not have to be uniform. In the interest of keeping this discussion simple, we elect to make uniform divisions. Suppose we wish to make uniform divisions of the real line, starting at 0 of a selected size $\Delta x$. From zero, in both directions, we count out the edges of our bins by increments of $\Delta x$. Since the probability calculations will require a starting point, we need to choose a location to represent each of these intervals. We use the midpoints of these intervals to represent the countable state space. In Figure M.5, these states are illustrated by circles.

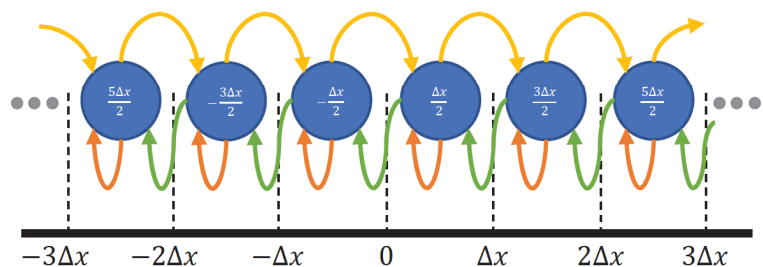

Figure M.5: Illustration on the creation of a Markov chain on the real line.

M-222    **M.2.1.2    Neighbors for each state in the Markov chain.**

M-223    We must now identify the neighbors of each state in our space. As previously discussed, given a
M-224    starting location, the process $X(t)$ has a non-zero probability of transitioning to any of our defined
M-225    intervals on the real line. If we were to mimic this, then each state in our state space would be a
M-226    neighbor to all other states. That is, the graph representing the possible transitions between state
M-227    spaces would be a complete graph.

M-228    However, neuromorphic hardware and practical limitations (i.e., fan-in/fan-out considerations)
M-229    suggest that we cannot allow each state space to transition to all others. This means a choice must
M-230    be made. In the context of neuromorphic hardware, we must decide how much fan-out we want.
M-231    In the examples in the text, we typically allowed transitions between adjacent states and back to
M-232    the original state. For this discussion, we allow transitions to the left and right, and back to itself
M-233    (see arrows in Fig. M.5).

M-234    **M.2.1.3    Calculation of transition probabilities.**

M-235    Our first two steps in the calculation of the Markov chain approximation involved two independent
M-236    choices: a choice of $\Delta x$ and a choice of neighboring states. To calculate the probabilities, a depen-
M-237    dent choice will be made based on the previous selections. Namely, we must select an appropriate
M-238    time discretization size $\Delta t$ for the discrete time Markov chain.

M-239    Again, recalling that (M.2) can transition to any interval on the real line given a starting
M-240    position, we must choose $\Delta t$ small enough so that the probability of transitioning to intervals
M-241    outside of the defined neighbors is smaller than some threshold probability. For all of our examples
M-242    in the text, we choose $\Delta t$ small enough so that transition outside of the chosen neighbors is less
M-243    than 0.05.

M-244    However, the choice of $\Delta t$ may also depend on another factor. If there is a Poisson process, that
M-245    is if $dP(t; Q, X(t), t)$ appears in the SDE, then we also want to choose $\Delta t$ small enough so that we
M-246    can be reasonably sure that *at most* one Poisson event can occur in any time window. Again, we
M-247    ensure $\Delta t$ is selected so that the probability of more than one Poisson event occurring in any time
M-248    window is less than 0.05.

M-249    To that end suppose that $x_i$ and $x_j$ are nodes and that $x_j$ is a neighbor of $x_i$. What is the
M-250    probability that $x_i$ transitions to $x_j$? For simplicity, we will assume that $h$ is deterministic. Then,
M-251    using (M.2), the probability of $x_i$ transitioning to $x_j$ in the time interval $[t, t + \Delta t]$ is the sum of
M-252    two probabilities. Given $X(t) = x_i$, the first is the probability that $X(t + \Delta t)$ is in the interval
M-253    represented by $x_j$ and that no Poisson jump occurs. The second is the probability that $X(t + \Delta t)$
M-254    is in the interval represented by $x_j$ and some Poisson jump occurred.

M-255    To help calculate these probabilities, we will appeal to the Euler-Maruyama simulation method
M-256    for SDEs. This is a discretization method for simulating sample paths of SDEs. The method
M-257    works similar to the Euler method in that the value of $X(t)$ is assumed constant over some interval
M-258    $[t, t + \Delta t]$ and an increment is calculated from this assumption. For (M.2), given $X(t) = x_i$ and
M-259    assuming that $h$ is deterministic, the Euler-Maruyama method gives

$$X(t + \Delta t) \approx x_i + b(t, x_i)\,\Delta t + a(t, x_i)\,W(\Delta t) + h(t, x_i)\,\mathbb{1}_{P(t;Q,x_i,t)}, \tag{M.3}$$

where $\mathbb{1}_{P(t;Q,x_i,t)}$ represents the indicator function of whether the Poisson process $P$ fired in the time window $[t, t + \Delta t]$ given that $X(t) = x_i$, and $W(\Delta t)$ is a normal random variable with mean 0 and variance $\Delta t$. Put another way, if the Poisson process *does not fire*, then $X(t + \Delta t)$ is assumed to be a normal random variable with mean $x_i + b(t, x_i)\Delta t$ and variance $a^2(t, x_i)\Delta t$. If the Poisson process *does fire*, then $X(t + \Delta t)$ is assumed to be a normal random variable with mean $x_i + b(t, x_i)\Delta t + h(t, x_i)$ and variance $a^2(t, x_i)\Delta t$.

Phrased in this way, calculation of the appropriate probabilities is clearer. Let

$$p_J(x_i, t, \Delta t) = \left(\int_t^{t+\Delta t} \lambda(s, x_i)\, ds\right) \exp\left(-\int_t^{t+\Delta t} \lambda(s, x_i)\, ds\right). \quad (\text{M.4})$$

This is the probability of a single Poisson event occurring in the window $[t, t+\Delta t]$ given that $X(t) = x_i$ is constant on the time interval. Let $\mathcal{N}(\mu, \sigma^2)$ denote a normal random variable with mean $\mu$ and variance $\sigma^2$. Then $p_{ij}$, the probability of transitioning from $x_i$ to $x_j$, is given by

$$\begin{aligned}p_{ij}(t) &:= \mathbb{P}\left[x_i \to x_j \mid t \to t + \Delta t\right] \\ &= (1 - p_J(x_i, t, \Delta t))\, \mathbb{P}\left[\mathcal{N}\left(x_i + b(t, x_i)\Delta t,\ a^2(t, x_i)\Delta t\right) \in \left(x_j - \frac{\Delta x}{2}, x_j + \frac{\Delta x}{2}\right)\right] \\ &\quad + p_J(x_i, t, \Delta t)\, \mathbb{P}\left[\mathcal{N}\left(x_i + b(t, x_i)\Delta t + h(t, x_i),\ a^2(t, x_i)\Delta t\right) \in \left(x_j - \frac{\Delta x}{2}, x_j + \frac{\Delta x}{2}\right)\right],\end{aligned} \quad (\text{M.5})$$

where $\mathbb{P}[\cdot]$ denotes the probability of the interior event occurring. This is a probabilistic representation of the intuition we discussed: the probability of transitioning is the sum of two probabilities, one of ending up in the interval with no jump and one of ending up in the interval having experienced a jump.

In the the event that the jump $h$ is random, a similar construction can be made taking care to sum or integrate over the distribution of rewards. Our steady-state particle transport problem is an example of this concept.

Letting $t_j = j\Delta t$, (M.4) can be used along with the probability of no jumps occurring

$$p_0(x_i, t_j, \Delta t) = \exp\left(-\int_{t_j}^{t_j+\Delta t} \lambda(s, x_i)\, ds\right)$$

to ensure that the probability of more than one jump occurring $(1 - p_0 - p_J)$ is less than 0.05 for all $x_i$ and $t_j$. Similarly, (M.5) can be used to ensure the probability of transitioning to neighbors other than the allowed transitions is less than 0.05 for all $x_i$. The time discretization $\Delta t$ should be chosen to satisfy both of these constraints.

Using (M.5), a transition tensor $C(t_\ell) = (p_{ij}(t_\ell))$ can be constructed representing the Markov chain. For each fixed $t_\ell$, we must conserve probability – the probability of transitioning from state $x_i$ to some other state must sum to 1. In the Markov chain, this means that $\sum_j p_{ij}(t_\ell) = 1$. In this discussion example, we are only allowing transitions to the left, right, and back to the same node. However, the true process $X(t)$ may transition to any interval on the real line; calculating the transition probabilities for just a subset of the possible transitions may mean that these probabilities do not sum to 1. While $\Delta t$ was chosen so that this probability of transition is small, we must require a sum of 1 to ensure the Markov chain is well defined.

There are various ways to conserve probability at this point. One possibility is to just normalize the rows. Another possibility is to add the missing amount to one of the transitions. Unless stated otherwise in our example discussions, we calculate as follows. If, in our example, we were calculating the transition of probability to the left node, we would calculate the probability of transitioning to the left node *or beyond*. Similarly for the right node. Again, since $\Delta t$ has been chosen so that transitioning more than a single space to the left or right is small, the error accrued from this assignment is also small.

It is important to call out that $C$ **does depend** on the time step $t_\ell$ whenever $a$, $b$, $\lambda$, or $h$ depend on $t$. When these functions do not depend on time, a static transition matrix is created. If time dependence exists and if a maximum desired time is known, the transition tensor $C$ can be collapsed into a single matrix. For example, if there are 10 possible state spaces and it is only desired to simulate for 100 time steps, then the one hundred $10 \times 10$ transition matrices for each time step would become a single transition matrix over a state space of size 1000.

Before discussing the restriction to a finite state space, we call out an implicit assumption made in this construction: namely that when a random walk transitions to an interval represented by the midpoint $x_j$, we assume that the walker takes on the value of the midpoint. This introduces some rounding error, discussed further below.

### M.2.1.4  Restriction to a finite state space.

For some problems, like our heat transport on the sphere or time-dependent particle transport examples, the construction above yields a finite state space. However, situations may arise where the method we have described yields a countably infinite state space. When considering hardware limitations, like the finite number of nodes on a spiking neuromorphic platform, it may be necessary to reduce to a finite state space.

Continuing our discussion of (M.2) on the real line, we would need to select a finite subset of our states to transition between. Due to the construction of neighbors, we probably would select a minimum interval and a maximum interval, keeping all states between. Without loss of generality, suppose that the states are ordered and that the minimum state corresponds to $x_1$ and the maximum state corresponds to $x_N$.

For each $t_\ell$, we pluck out the $N \times N$ section of the transition matrix $C(t_\ell)$ corresponding to our truncated state space. To conserve probability, for each state we add the total probability of a transition to a state less than $x_1$ to the probability of transitioning to $x_1$. Similarly for a transition to a state greater than $x_N$. In this example where transitions are restricted to the left, right, and to the same state, this addition of probabilities only occurs on the endpoints.

Once the transition matrix is finalized, it can be used to sample some number $M$ of random walks. If $x_j$ is some node in the state space, then we write $X_{j,i}(k\Delta t)$ for the location of the $i^{\text{th}}$ random walk at time $k\Delta t$ that started at position $x_j$.

M-323    ## M.2.2    Utilizing Sampled Random Walks to Approximate PIDE Solution

Consider the probabilistic solution from Theorem S2.1.1 in Supplementary Note 2:

$$\begin{aligned} u(t,x) = {} & \mathbb{E}\left[g(X(t))\exp\left(\int_0^t c(s,X(s))\,ds\right)\bigg| X(0)=x\right] \\ & + \mathbb{E}\left[\int_0^t f(s,X(s))\exp\left(\int_0^s c(\ell,X(\ell))\,d\ell\right)ds \bigg| X(0)=x\right]. \end{aligned} \quad (M.6)$$

We would like to evaluate $u(t_i, x_j)$ via this expectation using the Monte Carlo method and the sampled random walks for each position $x_j$ in the mesh and time point $t_i$. Since our random walks occur over discrete time points and occupy discrete locations, we do this via Riemann sum approximations for the integrals. If there were $M$ total random walks sampled that started on position $x_j$, then

$$\begin{aligned} u(t_i, x_j) \approx {} & \frac{1}{M}\sum_{\ell=1}^{M}\Bigg[g(X_{j,\ell}(i\Delta t))\exp\left(\sum_{k=0}^{i} c(t_k, X_{j,\ell}(k\Delta t))\Delta t\right) \\ & + \sum_{k=0}^{i} f(k\Delta t, X_{j,\ell}(k\Delta t))\exp\left(\sum_{s=0}^{k} c(s\Delta t, X_{j,\ell}(s\Delta t))\Delta t\right)\Delta t\Bigg]. \end{aligned} \quad (M.7)$$

M-324    ## M.2.3    Brief Commentary on Accuracy of Approximation

Here, we evaluate the implications of using a discrete spatial mesh for approximating random walks that ideally would be continuous valued. Such approximations are implicit in any numerical implementation of random walks on a system with finite precision, although since our implementation directly implements a discrete mesh to describe the state-space, it is necessary to consider the numerical implications. We approximate a jump-diffusion process with a discrete-time Markov chain. Weak convergence results for Markov chains converging to jump diffusions were found by Skorokhod [16]. Analytic, rather than probabilistic, proofs were later considered, determining an order of convergence of $\mathcal{O}(1/\sqrt{n})$ [10].

Fixing some initial condition and setting $\Delta x = 1/n$, let $X_n(t)$ represent the Markov chain approximation of (M.2). By saying that $X_n(t)$ converges weakly to $X(t)$, we mean that as $n \to \infty$, or rather as $\Delta x \to 0$, the transition density of $X_n$ converges to that of $X$. If a function $\Psi$ is continuous and bounded and if $X_n$ converges weakly to $X$, then $\mathbb{E}^*[\Psi(X_n)]$ converges weakly to $\mathbb{E}[\Psi(X)]$, where $\mathbb{E}^*$ represents the outer expectation (or outer measure). This would imply that, under appropriate conditions, we have convergence of (M.7) to (M.6) as we decrease $\Delta x$. Beyond the scope of this work, error bounds for continuous time Markov chain approximations ($\Delta t \to 0$) have been found for PIDEs involving fractional time operators [9].

Setting aside these concepts of order of convergence, we would like to demonstrate an error estimate on using the discretization of space in the *best case scenario*. To that end, fix some time $t$ and a position $x$. Suppose we wish to approximate

$$u(t,x) = \mathbb{E}[\Psi(X(t))\,|\,X_0 = x]$$

M-344  by sampling $M$ paths of $X(t)$. Let $X_i$ be the $i^{\text{th}}$ sampled path. Then the Monte Carlo approximate
M-345  solution is

$$u_M(t,x) = \frac{1}{M}\sum_{i=1}^{M}\Psi\left(X_i(t)\right).$$

M-346  To mimic the discrete spatial approximation in the best case scenario, let's assume that there
M-347  exist true random paths $\Psi(X_i(t))$ and that our approximation "snaps" or rounds the true value of
M-348  $\Psi(X_i(t))$ to the nearest point on a spatial grid with size $\Delta x$. We will denote this rounded process
M-349  by $\widehat{\Psi}(X_i(t))$ and the resulting approximate Monte Carlo solution by

$$\widehat{u}_M(t,x) = \frac{1}{M}\sum_{i=1}^{M}\widehat{\Psi}\left(X_i(t)\right).$$

In this manner, at time $t$, the process $\widehat{\Psi}(X_i(t))$ is at most $\Delta x/2$ away from the true sample path $\Psi(X_i(t))$. The error of this best case approximation is estimated as

$$|u(t,x) - \widehat{u}_M(t,x)| \leq |u(t,x) - u_M(t,x)| + |u_M(t,x) - \widehat{u}_M(t,x)|$$

$$\leq |u(t,x) - u_M(t,x)| + \frac{1}{M}\sum_{i=1}^{M}\left|\Psi(X_i(t)) - \widehat{\Psi}(X_i(t))\right|$$

$$\leq |u(t,x) - u_M(t,x)| + \frac{\Delta x}{2}.$$

M-350  The first term on the right hand side is the Monte Carlo error and decays as $1/\sqrt{M}$. The second
M-351  term is the error accrued due to forcing our process $\Psi(X(t))$ to snap to a grid. Notably, it cannot
M-352  be controlled by the number of samples $M$ and can only be made small with a sufficiently small
M-353  grid.
M-354       When using the Markov chain approximation, however, we do not have the actual path and
M-355  instead calculate the next value in the path by assuming the random walk is in the center of a
M-356  voxel and then travels to the center of another voxel with a probability determined from landing
M-357  anywhere in the voxel. The error that arises by this method would increase on each time step.
M-358  Even in the best case scenario, the error accrued on the $j^{\text{th}}$ time step would be dependent on the
M-359  function $\Psi$ applied to the underlying spatial grid. If $\Psi$ is the identity, and we are snapping a true
M-360  sampled path $X_i(t)$ to a grid, then the error accrued in an absolute best case scenario for the $j^{\text{th}}$
M-361  time step would be on the order $j\Delta t\Delta x/2$.
M-362       Reducing to a finite state space is problem specific. The error heavily depends on the application
M-363  and how the finite states are selected. Additionally, the examples we consider do not require this
M-364  truncation. As such, we will not explore the consequences of a finite state space here.

# Supplementary Note 1: Complexity, Scaling, and the Neuromorphic Advantage

This note is primarily concerned with discussing a neuromorphic advantage by analyzing the computational complexity of our random walk algorithm across conventional and neuromorphic platforms (Section S1.1). We perform additional TrueNorth studies in Section S1.2. We close this note with a small discussion on the hardware differences between Loihi and TrueNorth in Section S1.3.

## S1.1   Computational Complexity

There is no fundamental reason to expect neuromorphic computing to belong to a different computational complexity class (such as allowing exponential algorithms to be computed in polynomial time; as can be the case with quantum computing) than von Neumann architectures, as proposed neuromorphic systems largely leverage the same physics. However, several aspects of neuromorphic computing can be formally shown to provide scaling benefits. Specifically, neuromorphic hardware can be shown to provide formal advantages due to both reduced communication inherent in processing-in-memory [2], and high neuron fan-in/fan-out [15, 12].

In this paper, we define a neuromorphic advantage as an algorithm that shows a demonstrable advantage in one resource (e.g., energy) while exhibiting comparable scaling in other resources (e.g., time).

Unfortunately, making apples-to-apples comparisons of an NMC algorithm to a VN algorithm is non-trivial, because while the objective may be the same, the respective algorithms should be tailored to the underlying architecture (stated differently, our spiking algorithm would be an inefficient approach to compute random walks on standard hardware, and neuromorphic hardware cannot directly implement a conventional implementation). For this reason, we compare our NMC algorithm to the most straightforward VN algorithm, even though these implementations may be formulated quite differently. This analysis is relatively straightforward for the case of a simple Markovian random walk, which we explore here.

### S1.1.1   Objective

For this analysis, we consider a simple Markovian random walk model over a pre-defined state space of size $K$, which for most applications we will benefit from increasing the number of walkers to the greatest extent permitted by computational resources. While the approach described in our paper can extend to more complex physics, such as particle absorption, we focus on the simplest scenario here.

For our analysis, we will consider the time ($T$) and energy ($E$) scaling of independently simulating a number of walkers, $W$, over $S$ time steps.

To simplify the analysis, we focus our analysis on a hypothetical single-chip system, though the analysis extends to multi-chip systems insofar as chip to chip communication is similar between architectures. We further focus exclusively on the DTMC simulation itself, as opposed to any averaging or other subsequent analysis of the outputs of the runs.

### S1.1.2 Complexity of Random Walks on a Parallel von Neumann System

We consider specifically the case of a system with multiple von Neumann processors in parallel, such as a CPU with multiple cores. We define $P$ as the number of processors, or programmable cores, on the chip. In the absence of interactions, the most efficient straightforward implementation of a large number of walkers would be to distribute the walkers evenly over $P$ and walkers would persist on that core through the length of the simulation. Under these conditions, the time of the simulation, $T_{\text{VN}}$, would be

$$T_{\text{VN}} = C_{\text{VN, time}} \frac{W \times S}{P}, \tag{S1.1}$$

where $C_{\text{VN, time}}$ is a constant describing the hardware-dependent time cost of the RW operations on the von Neumann architecture. Likewise, the total energy required to perform the DTMC simulation would be

$$E_{\text{VN}} = C_{\text{VN, energy}} \left( W \times S \right), \tag{S1.2}$$

where $C_{\text{VN, energy}}$ is a constant describing the hardware-dependent energy cost of the RW operations.

From these equations, it can be seen that increasing the density of cores on a chip (or number of chips) can speed up a simulation dramatically, however it does not yield any power savings, which remains proportional to the total computational work of the task.

### S1.1.3 Complexity of Random Walks on NMC

The NMC algorithm that we consider is fundamentally different than the straightforward VN algorithm. While it is possible to dedicate a subset of neurons to model each walker (see the particle method in [14]), the approach explored here uses neurons to explicitly model the state space over which walkers may randomly walk as a graph, with a small circuit of neurons at each mesh point. At each simulation time, the circuit at each mesh point distributes its walkers through its edges by the probabilities defined in the Markovian transition matrix. This method, which is referred to as the density method in [14] since the algorithm directly represents the density of walkers at each time step, has a time cost given by

$$T_{\text{neural}} = c_{\text{neural, time}} \frac{W \times S}{M}, \tag{S1.3}$$

where $c_{\text{neural, time}}$ is the corresponding time-cost of updating a single walker's position on the mesh for one time step.

More complicated, however, is the denominator $M$, which is the neuromorphic analogue to the number of cores, $P$, on a VN processor. $M$ both captures the inherent parallelism of neurons on a chip, which varies across architectures, as well as the distribution of walkers over the neurons, which is a non-deterministic characteristic.

In an ideal NMC system, where every neuron is computed truly in parallel, $M$ would approach a maximum value, $M_{\text{max}}$, which is the number of mesh-points $K$, in the state space. Currently, most neuromorphic architectures leverage cores that are responsible for computing a subset of neurons, making computation across cores fully parallel with differing degrees of parallelism within cores.

S1-71   So conservatively we can also limit $M_{\text{max}}$ to no greater than the number of distinct neural cores
S1-72   $N_{\text{cores}}$ on a chip.
S1-73        In addition, $M$ will vary with walker distribution over time. If walker updates are synchronized,
S1-74   then the DTMC simulation can only advance forward a time step after all walkers have been
S1-75   processed at each mesh node. If we consider an initial condition where all walkers start at the same
S1-76   mesh point, it will take $W$ hardware time steps for the model to progress one time step; however,
S1-77   if the model is fully mixed (walkers distributed evenly over the mesh), then the time to simulate
S1-78   one time step drops considerably.
S1-79        In practice, this will make $M$ highly dependent on both the physics being modeled (how fast
S1-80   do the walkers distribute over the mesh), the ratio of number of walkers to the mesh-size, and the
S1-81   relevant time of simulation (longer simulations permit more mixing and thus more time-efficient
S1-82   time steps). Further, in practice one could replicate mesh-points which can be anticipated as
S1-83   chokepoints due to initial conditions (such as parallelizing start nodes), but this is not always
S1-84   foreseeable in simulations.
S1-85        For our purposes, we consider that $M$ will eventually approach the minimum of the number of
S1-86   mesh points and number of independent neural cores for long simulations. That is,

$$M \approx \min\left(N_{\text{cores}}, K\right). \tag{S1.4}$$

S1-87   However, as with the conventional processing, the total energy consumption should be independent
S1-88   of $M$:

$$E_{\text{neural}} = c_{\text{neural, energy}} \left(W \times S\right). \tag{S1.5}$$

S1-89   ### S1.1.4   Identifying a Neuromorphic Advantage

S1-90   Per our definition above, our approach can show a neuromorphic advantage is if the time-scaling of
S1-91   the algorithm on neuromorphic hardware is preserved while showing an energy advantage. While
S1-92   we know neuromorphic hardware shows an absolute power advantage (typical neuromorphic chips
S1-93   have sub-1W power requirements vs 100W for server-class CPUs), a power advantage can be easily
S1-94   offset if the computation takes much longer to complete.
S1-95        We performed scaling studies on a single TrueNorth chip, a single Loihi chip from an 8-chip
S1-96   Nahuku board, a commodity class dual CPU system, using Intel Xeon E5-2665, which has 8 CPU
S1-97   cores capable of 16 threads, and an NVIDIA Titan-XP GPU (expanded GPU details are below).
S1-98   While some of these platforms are several years old at this point, they are representative of the state-
S1-99   of-the-art in process engineering at the time of their development. We programmed each platform
S1-100  to implement a simple random walk over a small mesh ($21 \times 21$ spatial grid, configured as a torus)
S1-101  using a platform-appropriate algorithm. The implementation was using C++ with OpenMP to
S1-102  leverage multiple threads on the Xeon CPUs, C++ with CUDA for the GPU, MATLAB corelets
S1-103  for the TrueNorth implementation, and NxNET using Python for Loihi.
S1-104       We measured scaling in two ways. First, we measured model scaling on a single core / NMC mesh
S1-105  by progressively increasing the number of walkers on a single core / mesh. Second, we performed
S1-106  a standard "weak scaling" experiment; increasing the number of walkers along with the number of
S1-107  cores / NMC meshes. For the GPU, we distributed the walkers onto a single thread block (1024

s1-108 threads) or allocated resources to match the number of walkers in the simulation (GPU-weak).
s1-109 Each simulation ran for 100,000 walker updates with walker counts increasing through [1000, 2000,
s1-110 4000, 8000, 12000, 16000, 24000, 32000]. Through the C++ code on the CPU and GPU, MATLAB
s1-111 scripts for TN, and Python scripts for Loihi we measured only the simulation time, ignoring all
s1-112 costs associated with model setup or post-processing of the walkers. We further eliminated I/O
s1-113 from the chips to the fullest extent possible, as depending on the system can dominate processing
s1-114 cost. For each of these simulations, we performed 10 replicate runs, though the variability across
s1-115 runs was typically negligible.

s1-116 From these scaling studies, we first measured the time to complete these simulations (raw data
s1-117 available in supplemental data file). As expected, the GPU implementation is considerably faster
s1-118 than the neuromorphic hardware (Figure S1.1), however the Loihi platform is nearly as fast as the
s1-119 CPU. However, we do see that the ability of a single NMC mesh to distribute additional walkers
s1-120 provides NMC with an advantage in terms of time-scaling, suggesting that a single NMC mesh
s1-121 scales at the rate of multiple CPU cores (Figure S1.2a). The weak scaling experiment tracks a
s1-122 different form of scaling, whereby we use multiple CPU cores / NMC meshes to simulate walkers
s1-123 in parallel. Here, we observe that both the CPU and NMC implementation scale similarly, at least
s1-124 up until the CPU starts using multi-threading on cores (Figure S1.2b). *Combined, these results*
s1-125 *confirm that the scaling of our NMC algorithm on NMC architectures scales similarly, and possibly*
s1-126 *slightly better, to the standard CPU algorithm on CPUs.*

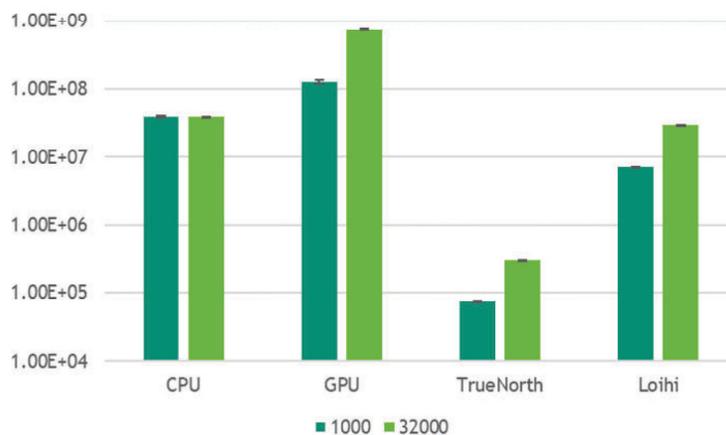

Figure S1.1: Walker updates per second for a 1000 (dark green) and 32000 (light green) basic diffusion simulation across conventional and neuromorphic platforms.

s1-127 From these time measurements, based on (S1.1) and (S1.3) above, we approximated the quan-
s1-128 tities $P/c_{\text{VN, time}}$ and $M/c_{\text{neural, time}}$, since the meaning of $M$ is poorly defined. In both cases,
s1-129 these quantities can be measured in units of ***walker updates per second***. Similarly, from the
s1-130 number of CPU and NMC cores used and published estimates of maximum power consumption of
s1-131 these chips ([1] and [11]), we can roughly estimate the power consumption in watts, or joules per
s1-132 second. Combining these two measures, we can then approximate the number of ***walker updates***

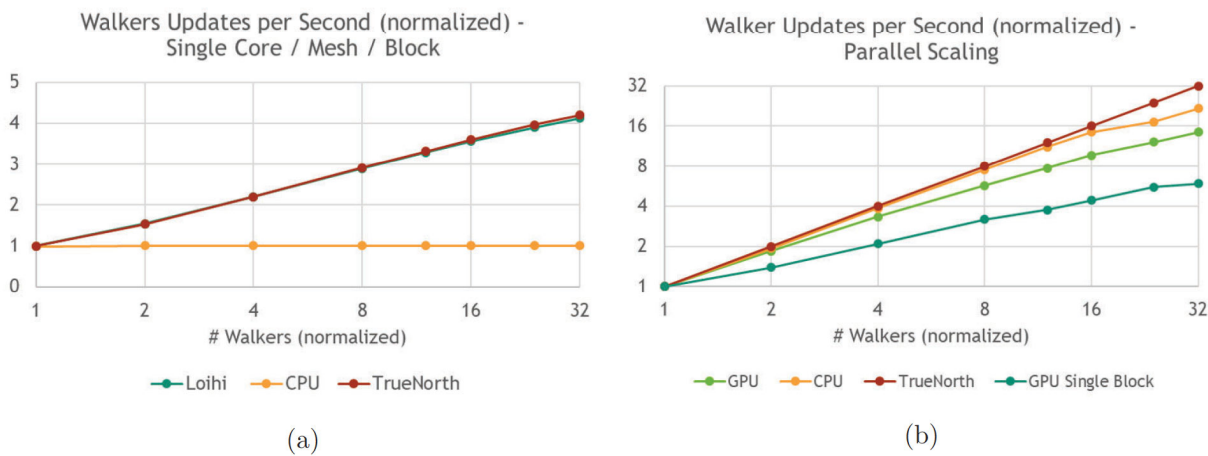

Figure S1.2: Normalized time comparison of a simple diffusion simulation accomplished on conventional and neuromorphic as a function of increasing random walkers. All times normalized to the time it takes to complete a simulation with 1000 walkers. Left: comparison of Loihi and single-chip TrueNorth to a single-core CPU simulation. Right: Comparison of multi-chip TrueNorth to multi-core CPU and GPU simulations. GPU generates threads for all walker scenarios; GPU Single Block allocates only 1024 threads for all walkers.

*per joule*.

*This energy-centered analysis shows that despite being considerably slower than conventional CPUs, NMC platforms can be considerably more energy-efficient than von Neumann processors.* Not surprisingly, regardless of on-chip parallelization, CPU energy scaling is relatively constant at between 2.5 and 3 million walker updates per Joule. Based on core-utilization estimates, the NMC is considerably more efficient and becomes more so at larger simulation sizes, starting at about 60 million walker updates per Joule and progressing up to as high as 250 million walker updates per Joule when walker density is high.

It is important to acknowledge that this is an indirect measure of power consumption. For instance, the CPU power measurement used above was based on linearly estimating power requirements based on core usage relative to published values of Total Draw Power (TDP) for a maximum load. Likewise, on event-driven NMC platforms, such as TrueNorth, the actual power draw should be a function of spiking activity, not core usage. It therefore would not be surprising to see these efficiency measures vary by several fold in either direction if direct measures were feasible. Nevertheless, the magnitude of the energy differences observed (between 10x and 100x) would show a considerable NMC advantage even if these estimates were off considerably.

## S1.2   Additional TrueNorth Scaling

Supplementing the previously described scaling studies (see Methods), we performed 3 additional studies evaluating the random walk algorithm on TrueNorth. In our third experiment, the total

S1-152  count of walkers is held fixed at 4000, but multiple copies of the mesh are instantiated, and the 4000
S1-153  walkers are equally distributed over all the copies. The fourth experiment is the same as our initial
S1-154  scaling study on a static mesh, however every node is assigned a random transition probability.
S1-155  These are drawn from a uniform distribution, and the sum of all transition probabilities is forced
S1-156  to sum to 1. The final experiment increases walkers by maintaining a single copy of the mesh, but
S1-157  distributing each additional set of 4000 random walkers over multiple starting locations, chosen at
S1-158  random from a uniform distribution.

S1-159  Experiment 3 further explores the nature of parallelizing the random walk by holding the number
S1-160  of walkers constant at 4000 and distributing them equally over more and more mesh copies. That
S1-161  is, by the last data point of Figure 2, at 50 copies of the mesh there are 80 walkers present on
S1-162  each mesh. As can be seen in figure 3 we achieve a significant reduction in execution time with
S1-163  diminishing returns beginning at 16 mesh copies. This is specific to the mesh size used in this
S1-164  experiment, however. With a $21 \times 21$ mesh size there are 441 nodes. Therefore, as the walkers
S1-165  spread out if there are less walkers than total nodes then the mean spread is tending towards one
S1-166  walker per node which is the lower bound for execution time in terms of the time step to tick ratio.
S1-167  At 16 mesh copies and 4000 walkers there are 250 walkers per mesh copy. Fewer walkers on the
S1-168  mesh will result in a quicker spread to the lower bound of execution time of the algorithm. The
S1-169  results of Experiment 3 are displayed graphically in Figure S1.3, while the data points are in our
S1-170  supplemental data file.

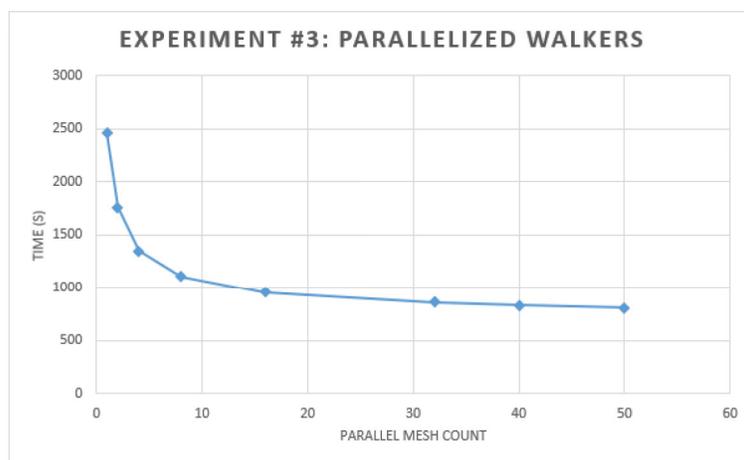

Figure S1.3: Results of TrueNorth Experiment #3. Execution time reaches a limit as mesh counts increase.

S1-171  Experiment 4 demonstrates the random walk algorithm's sensitivity to transition probabilities.
S1-172  In this experiment the transition probabilities between nodes where randomized. Figure 3 shows
S1-173  that we still achieve linear scaling as the number of walkers increase, but the error bars grow
S1-174  significantly as the mesh becomes crowded with walkers. The error bars are defined as 1 standard
S1-175  deviation of the experiment's trial set. These results show the sensitivity of the algorithm to
S1-176  execution time bottle necks due to transition probabilities that define areas of the mesh that have

S1-177 high likelihood of walkers entering but low likelihood of them leaving. Such a situation would cause
S1-178 the mean step ratio to be very high, lengthening the number of ticks required to reach 100,000
S1-179 simulation steps. The results of Experiment 4 are plotted in Figure S1.4 and data points are
S1-180 available in our supplemental data file.

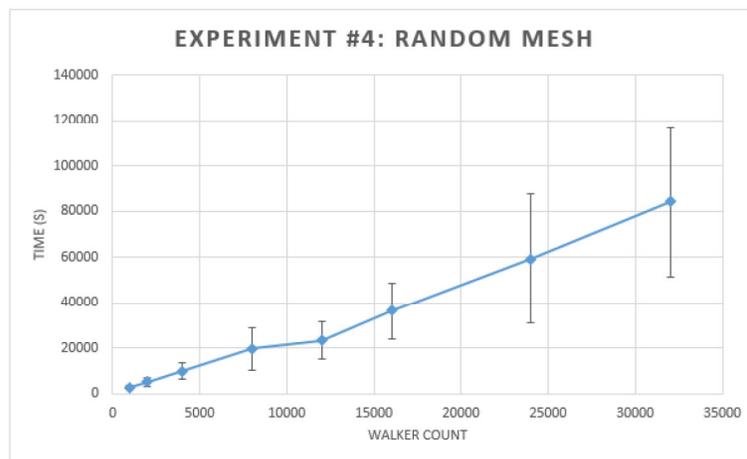

Figure S1.4: Results of TrueNorth Experiment #4. Execution time scales linearly with walker count, again, but also demonstrates the sensitivity of the algorithm to bottlenecks caused by uneven transition probabilities.

S1-181 Experiment 5 demonstrates the effect of providing an *a priori* spread of the walkers. From
S1-182 Figure 4 we see that a slight initial spread has a large effect on reducing execution time. We
S1-183 also reach diminishing returns very quickly. The initial data point is 4000 walkers all starting at
S1-184 one node on the mesh, which is consistent with all other experiments. The second data point is
S1-185 putting 1000 walkers each at 4 random locations and we see that any further initial spread does
S1-186 not make a significant difference in execution time. It is also interesting to point out that the
S1-187 standard deviation of the trial sets are noticeable, on the order of 10% of the mean. This is because
S1-188 some random choices could clump walker starting nodes together which would then have a greater
S1-189 likelihood of moving back and forth to each other, slowing the diffusion. Whereas starting locations
S1-190 that are maximally separated in the mesh will diffuse faster and thus execute faster. The results of
S1-191 this final experiment are displayed in Figure S1.5 and data are given in our supplemental data file.

## S1.3   Implications of TrueNorth and Loihi Neuromorphic Implementations

S1-194 As described in the Methods, the specific circuit implementations of our neural algorithm on
S1-195 TrueNorth and Loihi had subtle differences due to hardware-specific considerations such as neuron
S1-196 types and where random numbers are accessible within the circuit. While hardware-specific circuit
S1-197 alterations may introduce constant costs (e.g., a few extra neurons per mesh point), we expect that they should not affect the scaling of the algorithm, which we do observe in the generally comparable scaling between these two platforms. There are several considerations that are worth expanding on further, as they are generally applicable to NMC.

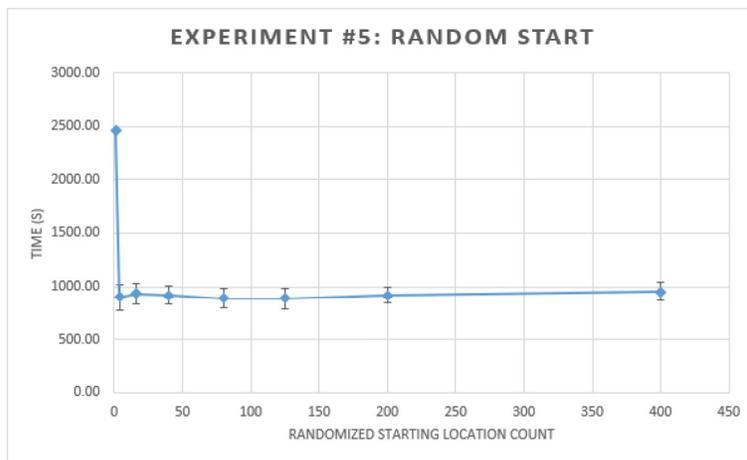

Figure S1.5: Results of TrueNorth Experiment #5. Execution time is dramatically reduced once all walkers do not start on the same position.

### S1.3.1   Hardware-specific Neural Circuit Partitioning

The first consideration is how to partition a neural circuit onto the neural processing cores on different NMC platforms. Both platforms consist of many effectively independent neural cores which are responsible for a subset of neurons. As with conventional parallel programming systems, while these cores operate in parallel with one another; within each core neural processing is not fully parallel, and communication between cores is often a dominant processing cost. As a result, strategies to partition neural circuits onto these cores will be an opportunity for future optimization, and these strategies will often be constrained by hardware-specific restrictions. Figure S1.6 illustrates how a simple mesh of our algorithm would be partitioned over 5 neural cores on Loihi and TrueNorth. One key difference shown in the figure is that the TrueNorth architecture allows different neuron types on the same core, whereas our Loihi NxNet implementation requires that neurons with different stochastic properties be placed on different cores, increasing inter-core communication costs between neurons within a single mesh of points. A separate consideration that arises in more complex implementations is that TrueNorth neurons can communicate at most to one other core, which requires that some neurons be duplicated if the algorithmic neuron's fan-out would need to target neurons on two different cores.

Importantly, the implications of this neural circuit partitioning is still an open research question and an opportunity for further optimization for algorithm and compiling research. For instance, as we consider larger scale meshes that would span multiple chips, on Loihi it will be important to develop embedding strategies that ensure deterministic and probabilistic neurons for mesh points

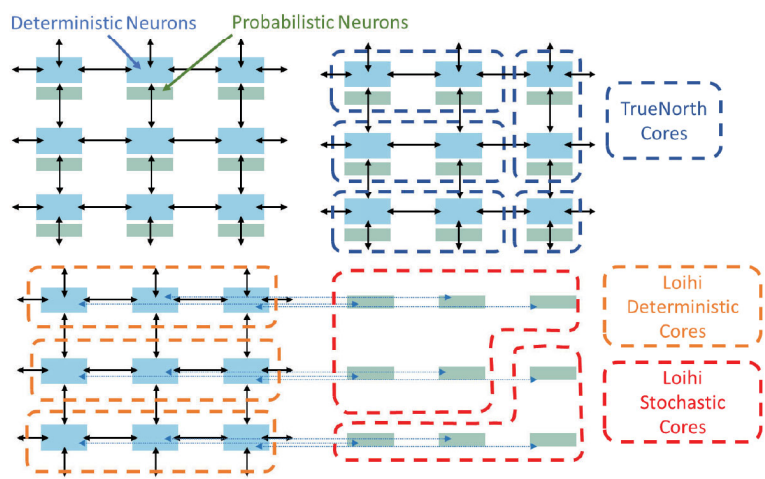

Figure S1.6: Illustration of the partitioning of our mesh onto Loihi and TrueNorth. Loihi requires deterministic and stochastic neurons be separated whereas TrueNorth allows for mixing.

stay on the same chip.

### S1.3.2   Spike Routing and Timing

The mechanism by which spikes are communicated between cores is one of the most variable features across neuromorphic platforms and has significant implications on the relative performance of these platforms. Some NMC architectures leverage fully asynchronous spike delivery with no guarantees of specific timing since a presumed benefit of brain-like algorithms on NMC is that biology-inspired algorithms can often be more tolerant of spikes arriving slightly earlier or later (or not at all). Unlike more brain-inspired algorithms, the algorithm we present in this paper requires very precise spike-timing and is very sensitive to any missing spikes. As such, this makes the algorithm not as immediately well-suited for NMC architectures which derive performance advantages from less precise timing in their information routing. While our algorithm can be ported to systems such as SpiNNaker whose spike routing is more asynchronous, it will require several mitigation strategies (custom neuron types; slowed system clock speeds) to ensure that information is not lost. On Loihi, the main implication of asynchronous spike delivery here is that we have to artificially build in a minimal delay in every neural connection to ensure that all spikes are delivered for a time step prior to neurons updating. One opportunity for future Loihi-specific optimization to mitigate this is to leverage Loihi's multi-compartment neurons to avoid these delays in our counter circuits. On TrueNorth, the main implication of its routing is that the chip has to be operated at a relatively slow clock-speed to avoid spike collisions in its custom core-to-core routing. As with the circuit embedding above, it is likely possible to optimize circuit layout on chip to reduce the risk of such communication bottlenecks.

## S2-1,S2-2 Supplementary Note 2: Connecting a Class of PIDEs with a Probabilistic Representation

S2-3 This note provides references and covers results concerning the probabilistic representation of solutions to the PIDEs considered in the main text.

S2-5 The underlying tools used here are the stochastic analogs of methods from differential calculus. Specifically, the main tool leveraged is Itō's rule (sometimes called Itō's lemma or Itō's formula). This landmark result from stochastic calculus is often referred to as the stochastic chain rule and serves the same purpose as the traditional chain rule in calculus. The two proofs in this note follow the same general road map. Itō's rule is used to determine a form for $\mathrm{d}u(t, X(t))$, where $X(t)$ is a stochastic process. Once this is found, integration in time followed by an expectation yields the final result.

S2-12 In the results provided below, we assume that a unique solution to the problem exists. Then, we show that the unique solution has a probabilistic representation. We do not address conditions for the existence and uniqueness of a solution. Classical solutions to these types of problems are discussed in [6, 18]. For this style of problem with no integral term, conditions for existence, uniqueness, and the probabilistic representation are covered in [7].

S2-17 The results for probabilistic representations were heavily influenced by techniques used in financial math. As such, the original results typically concern final value problems. Indeed, [18] contains results of this flavor. The primary source we use for the results shown is [8]. This text determines the Kolmogorov equations for a generic jump-diffusion process. Using these derivations, the author discusses a probabilistic representation for the final value problem of the PIDE considered in the main text.

S2-23 It is beyond the scope of this work to provide a full introduction to stochastic calculus. However, we provide a few short details of stochastic differential equations for intuition purposes. Consider the stochastic differential equation

$$\mathrm{d}X(t) = b(t, X(t))\,\mathrm{d}t + a(t, X(t))\,\mathrm{d}W(t) + h(t, X(t))\,\mathrm{d}P(t).$$

S2-26 In this equation, $W(t)$ is a Brownian motion and $P(t)$ is a Poisson process. Those with some background in white noise processes may feel uneasy with this equation since Brownian motion is nowhere differentiable. Therefore the concept of "$\mathrm{d}W(t)$" does not make much sense. However, this equation is merely shorthand for the integral equation

$$X(t) = X(0) + \int_0^t b(s, X(s))\,\mathrm{d}s + \int_0^t a(s, X(s))\,\mathrm{d}W(s) + \int_0^t h(s, X(s))\,\mathrm{d}P(s).$$

S2-30 In this context, $\mathrm{d}W(t)$ feels more like the concept of Riemann-Stieltjes integration. Indeed, integration against Brownian motion can be shown to make mathematical sense. Similarly, integration against a Poisson measure is well defined. The integrals with respect to Brownian motion and the Poisson process are called stochastic integrals. The expectation of the Brownian motion one is always zero. When the Poisson measure has mean zero, the expectation of the Poisson integral can also be shown to be zero.

Returning to the SDE, we are now ready to apply some intuition. The term $b(t, X(t))$ is the drift, or velocity, term. This describes the overall trend of the process $X(t)$. The next term, $a(t, X(t))$ is the diffusion term. This term describes the noise associated with the process $X(t)$. The final term $h(t, X(t))$ is a special term. This describes the value of any jumps in the process. The Poisson process $P(t)$ determines the times of the jumps. For further reading, [20] provides a basic overview of the stochastic calculus limited to Brownian motion; an in-depth approach covering stochastic integration against more general processes is given in [13].

We have organized this note as follows. In Section S2.1, we prove a representation theorem for the time-dependent initial value problem. In Section S2.2, we prove an analogous representation theorem for a special case steady-state boundary value problem. We adopt the notation from [8]. Hence, in all that follows, allow a semicolon in an argument to denote that the function might retain the variables to the right of the semicolon. An example of this retention occurs in our particle transport problems (see Supplementary Note 3). In these problems, the change in direction of transport after a scattering event depends on the previous direction. Hence, the previous direction appears in the probability density function.

## S2.1 A Probabilistic Solution for an Initial Value Problem PIDE

The following result is discussed for a final value problem in [8]. To obtain the result for an initial value problem considered, only a change of variables is needed. Rather than citing the result, we instead prove the result for the initial value problem with a slight alteration. Namely, we generalize the probability space so that the result can be directly applied to geometries like the sphere.

**Theorem S2.1.1.** *Let $(\Sigma \subseteq \mathbb{R}, \mathcal{F}, \mathbb{P})$ be a probability space that admits a Brownian motion $W(t)$ and suppose $x \in \Sigma$ and $t \in [0, \infty)$. Let $a, b, \lambda, c, f, u : [[0, \infty) \times \Sigma] \to \Sigma$ and $g : \Sigma \to \Sigma$ with $a \geq 0$. Denote by $P(t; Q, x)$ a Poisson process such that*

$$\mathbb{E}\left[\mathrm{d}P(t; Q, x, t)\right] = \lambda(t, x)\mathrm{d}t,$$

*where $Q$ is the jump-amplitude mark random variable of the process with co-domain in $\mathcal{Q}$ and probability density function $\phi_Q$. Let $h : [0, \infty) \times \Sigma \times \mathcal{Q}] \to \Sigma$. Define the stochastic process*

$$\mathrm{d}X(t) = b(t, X(t))\,\mathrm{d}t + a(t, X(t))\,\mathrm{d}W(t) + h(t, X(t), q)\,\mathrm{d}P(t; Q, X(t), t). \tag{S2.1}$$

*Consider the initial value problem*

$$\begin{aligned}
\frac{\partial}{\partial t}u(t, x) &= \frac{1}{2}a^2(t, x)\frac{\partial^2}{\partial x^2}u(t, x) + b(t, x)\frac{\partial}{\partial x}u(t, x) \\
&\quad + \lambda(t, x)\int \left(u(t, x + h(t, x, q)) - u(t, x)\right)\phi_Q(q; t, x)\,\mathrm{d}q \\
&\quad + c(t, x)u(t, x) + f(t, x),
\end{aligned} \tag{S2.2}$$

$$u(0, x) = g(x).$$

*Suppose:*

S2-59  - The functions $a$, $b$, and $h$ are continuously differentiable in all arguments and their spatial
S2-60    gradients are bounded.

S2-61  - The functions $c, f, g$ are bounded and continuous almost everywhere.

Then, if a unique solution to (S2.2) exists, it is given by

$$u(t,x) = \mathbb{E}\left[\exp\left(\int_0^t c(s, X(s))\,\mathrm{d}s\right) g(X(t)) \,\bigg|\, X(0) = x\right] \tag{S2.3}$$
$$+ \mathbb{E}\left[\int_0^t f(s, X(s)) \exp\left(\int_0^s c(\ell, X(\ell))\,\mathrm{d}\ell\right) \mathrm{d}s \,\bigg|\, X(0) = x\right].$$

S2-62

*Proof.* Let $u$ be the unique solution to (S2.2). Let some $t \in (0, \infty)$ be given. Let $s \in [0, t)$ represent the forward time (or time since zero) and $\tau = t - s$ represent the backward time (or time until $t$). Then $\widehat{u}(t - s, x) = u(s, x)$ satisfies the final value problem

$$-\frac{\partial}{\partial \tau}\widehat{u}(\tau, x) = \frac{1}{2}a^2(\tau, x)\frac{\partial^2}{\partial x^2}\widehat{u}(\tau, x) + b(\tau, x)\frac{\partial}{\partial x}\widehat{u}(\tau, x)$$
$$+ \lambda(\tau, x) \int (\widehat{u}(\tau, x + h(\tau, x, q)) - \widehat{u}(\tau, x))\phi_Q(q; \tau, x)\,\mathrm{d}q \tag{S2.4}$$
$$+ c(\tau, x)\widehat{u}(\tau, x) + f(\tau, x),$$
$$\widehat{u}(t, x) = u(0, x) = g(x).$$

S2-63  To simplify notation, denote

$$\widehat{u}_{\Delta h}(\tau, x, q) = \widehat{u}(\tau, x + h(\tau, x, q)) - \widehat{u}(\tau, x). \tag{S2.5}$$

For $\rho \in (\tau, t)$, consider the function

$$w(\tau, \rho, X(\rho)) = u(\rho, X(\rho))\, v(\tau, \rho),$$

where

$$v(\tau, \rho) = \exp\left(\int_\tau^\rho c(\ell, X(\ell))\,\mathrm{d}\ell\right).$$

Itō's rule combined with the stochastic product rule yield

$$\mathrm{d}w(\tau, \rho, X(\rho)) = v(\tau, \rho)\left[\left(\frac{\partial \widehat{u}}{\partial \rho}(\rho, X(\rho)) + b(\rho, X(\rho))\frac{\partial \widehat{u}}{\partial x}(\rho, X(\rho))\right.\right.$$
$$+ \frac{1}{2}a^2(\rho, X(\rho))\frac{\partial^2 \widehat{u}}{\partial x^2}(\rho, X(\rho))$$
$$\left.+ c(\rho, X(\rho))\, u(\rho, X(\rho))\right)\mathrm{d}\rho \tag{S2.6}$$
$$+ b(\rho, X(\rho))\frac{\partial \widehat{u}}{\partial x}(\rho, X(\rho))\,\mathrm{d}W(\rho)$$
$$\left.+ \int_{\mathcal{Q}} \widehat{u}_{\Delta h}(\rho, X(\rho), q)\,\mathcal{P}(\mathrm{d}\rho, \mathrm{d}q; X(\rho), \rho)\right],$$

where $\mathcal{P}(d\rho, dq; X(\rho), t\rho)$ is the Poisson random measure for the Poisson process; i.e.

$$\int_{\mathcal{Q}} \mathcal{P}(d\rho, dq; X(\rho), \rho) \equiv dP(\rho; \mathcal{Q}, X(\rho), \rho).$$

We rewrite the Poisson random measure as the sum of a mean-zero Poisson random measure and the mean:

$$\mathcal{P}(d\rho, dq; X(\rho), \rho) = \widehat{\mathcal{P}}(d\rho, dq; X(\rho), \rho) + \lambda(\rho, X(\rho))\,\phi_{\mathcal{Q}}(q; X(\rho), \rho)\,dq d\rho.$$

Using this in (S2.6) and appealing to (S2.4) yields

$$dw(\tau, \rho, X(\rho)) = v(\tau, \rho)\left[-f(\rho, X(\rho))\,d\rho + b(\rho, X(\rho))\frac{\partial \widehat{u}}{\partial x}(\rho, X(\rho))\,dW(\rho)\right.$$
$$\left. + \int_{\mathcal{Q}} \widehat{u}_{\Delta h}(\rho, X(\rho), q)\,\widehat{\mathcal{P}}(d\rho, dq; X(\rho), \rho)\right]$$

We now integrate both sides of this equation for $\rho \in (\tau, t)$ and then take an expectation. Since one is against Brownian motion and the other against a mean-zero Poisson measure, the expectation of the stochastic integrals are zero, yielding

$$\mathbb{E}\left[w(\tau, t, X(t)) - w(\tau, \tau, X(\tau))\,|\,X(\tau) = x\right] = \mathbb{E}\left[-\int_{\tau}^{t} v(\tau, \rho)f(\rho, X(\rho))\,d\rho\,\Big|\,X(\tau) = x\right].$$

The left-hand side of this equation becomes

$$\mathbb{E}\left[w(\tau, t, X(t)) - w(\tau, \tau, X(\tau))\,|\,X(\tau) = x\right] = \mathbb{E}\left[\widehat{u}(t, X(t))\exp\left(\int_{\tau}^{t} c(\ell, X(\ell))\,d\ell\right)\,\Big|\,X(\tau) = x\right]$$
$$- \widehat{u}(\tau, x).$$

Hence, we may rearrange and use the final condition to write

$$\widehat{u}(\tau, x) = \mathbb{E}\left[\widehat{u}(t, X(t))\exp\left(\int_{\tau}^{t} c(\ell, X(\ell))\,d\ell\right)\,\Big|\,X(\tau) = x\right]$$
$$+ \mathbb{E}\left[\int_{\tau}^{t} \exp\left(\int_{\tau}^{\ell} c(\rho, X(\rho))\,d\rho\right) f(\ell, X(\ell))\,d\ell\,\Big|\,X(\tau) = x\right]$$
$$= \mathbb{E}\left[g(X(t))\exp\left(\int_{\tau}^{t} c(\ell, X(\ell))\,d\ell\right)\,\Big|\,X(\tau) = x\right]$$
$$+ \mathbb{E}\left[\int_{\tau}^{t} \exp\left(\int_{\tau}^{\ell} c(\rho, X(\rho))\,d\rho\right) f(\ell, X(\ell))\,d\ell\,\Big|\,X(\tau) = x\right]$$

S2-64    Finally, we set $\tau = 0$ to yield (S2.3).                                                              □

In the main text, we mention that the function $c$ can be interpreted as an absorption term. We provide a little intuition for this declaration here. Consider (S2.2) with $\lambda, h, f$ all zero and assume $c(t,x) = -k$ for some $k \in \mathbb{R}^+$. Then, (S2.1) becomes

$$\mathrm{d}X(t) = b(t, X(t))\mathrm{d}t + a(t, X(t))\mathrm{d}W(t), \tag{S2.7}$$

and (S2.3) becomes

$$u(t,x) = \mathbb{E}\left[e^{-kt} g\left(X(t)\right) \,\Big|\, X(0) = x\right]. \tag{S2.8}$$

Imagine that we randomly kill the process (S2.7) according to a Poisson process with parameter $kt$. Then, the survival probability, or the probability that $X(t)$ has not been killed by time $t$ is the probability that no events have occurred by time $t$, or $e^{-kt}$. Then, we read (S2.8) as the average of $g(X(t))$ weighted by the probability that $X(t)$ has not died by time $t$. Rather than doing the weighting in the averaging, we could push this probability into the process $X(t)$ itself. We do this by terminating the process $X(t)$ according to the Poisson process described.

We finally note that Theorem S2.1.1 can be extended to multiple dimensions, but it requires significantly more bookkeeping. The extension to higher dimensions allows for the inclusion of multiple jump processes and allows for some correlation between stochastic processes. For examples on higher dimensions and further reading on these Feynman-Kac style results, see [7, 8, 4].

## S2.2   A Probabilistic Solution for a Boundary Value Problem Steady-State PIDE

Here we prove a probabilistic representation theorem for a special case of the previous theorem. This ordinary integro-differential equation is a steady-state boundary value problem where $c = 0$. A multiple dimension version of this theorem without an integral term is considered in [7]. Before proving the following theorem, we will discuss how this result intuitively follows the previous one.

**Theorem S2.2.1.** *Let $(\Sigma \subseteq \mathbb{R}, \mathcal{F}, \mathbb{P})$ be a probability space that admits a Brownian motion $W(t)$ and suppose $x \in D \subseteq \Sigma$. Let $a, b, \lambda, c, f, g, u : D \to D$ with $a \geq 0$. Denote by $P(t; Q, x)$ a Poisson process such that*
$$\mathbb{E}\left[\mathrm{d}P(t; Q, x, t)\right] = \lambda(x)\mathrm{d}t,$$
*where $Q$ is the jump-amplitude mark random variable of the process with co-domain in $\mathcal{Q}$ and probability density function $\phi_Q$. Let $h : [D \times \mathcal{Q}] \to D$. Define the stochastic process*
$$\mathrm{d}X(t) = b\left(X(t)\right)\mathrm{d}t + a\left(X(t)\right)\mathrm{d}W(t) + h\left(X(t), q\right)\mathrm{d}P\left(t; Q, X(t), t\right), \tag{S2.9}$$
*and the associated stopping time*
$$T_x = \inf\left\{t > 0 \,\middle|\, X(t) \notin D, X(0) = x\right\}. \tag{S2.10}$$
*Consider the boundary value problem*
$$\begin{aligned}
0 =\ & \frac{1}{2}a^2(x)\frac{\mathrm{d}^2}{\mathrm{d}x^2}u(x) + b(x)\frac{\mathrm{d}}{\mathrm{d}x}u(x) \\
& + \lambda(x)\int\left(u(x + h(x, q)) - u(x)\right)\phi_Q(q; t, x)\,\mathrm{d}q \\
& + f(x), \\
u(x) =\ & g(x), \qquad x \in \partial D.
\end{aligned} \tag{S2.11}$$

*Suppose:*

- *The functions $a$, $b$, and $h$ are continuously differentiable in all arguments and their spatial gradients are bounded.*

- *The functions $f, g$ are bounded and continuous almost everywhere.*

- *For any $x \in D$, $\mathbb{E}\left[T_x\right] < \infty$.*

*Then, if a unique solution to (S2.11) exists, it is given by*
$$u(x) = \mathbb{E}\left[g\left(X\left(T_x\right)\right) + \int_0^{T_x} f\left(X(s)\right)\,\mathrm{d}s \,\middle|\, X(0) = x\right]. \tag{S2.12}$$

To discuss how this result might intuitively follow from the first, consider (S2.2). If we wanted to consider a steady-state version, we could integrate in time, taking $t$ to infinity. This would effectively give us an ordinary integro-differential equation in $x$ for a time integrated version of $u$. In terms of the SDE, this would mean we would need to sample $X(t)$ for an arbitrarily long time.

If we then move from an equation on the whole plane to an equation on a more restrictive domain, we would need to somehow ensure that our process $X(t)$ does not travel outside of the domain of interest due to random fluctuations. We could accomplish this by artificially cutting off any sample of the process $X(t)$ when it first exits the domain of interest.

We now see how one might intuitively navigate from the time-dependent result to this one. Now we prove the result to fill in the details.

*Proof (of Theorem S2.2.1).* Let $u$ be the unique solution to (S2.11) and let $u_{\Delta h}$ be as defined in (S2.5) without the time argument. Let $n \in \mathbb{N}$ be given. Denote $T_x \wedge n = \min\{T_x, n\}$. By Itō's rule, for $t \leq T_x \wedge n$ we have

$$\begin{aligned} \mathrm{d}u\left(X(t)\right) &= \left(b\left(X(t)\right)\frac{\mathrm{d}}{\mathrm{d}x}u\left(X(t)\right) + \frac{1}{2}a^2\left(X(t)\right)\frac{\mathrm{d}^2}{\mathrm{d}x^2}u\left(X(t)\right)\right)\mathrm{d}t \\ &\quad + b\left(X(t)\right)\frac{\mathrm{d}}{\mathrm{d}x}u\left(X(t)\right)\mathrm{d}W(t) + \int_{\mathcal{Q}} u_{\Delta h}\left(X(t),q\right)\mathcal{P}\left(\mathrm{d}t,\mathrm{d}q;X(t),t\right). \end{aligned} \quad (\text{S2.13})$$

Similar to the previous problem, we rewrite the Poisson random measure as the sum of a mean-zero Poisson random measure $\widehat{\mathcal{P}}$ and the mean. We then integrate in time from 0 to $T_x \wedge n$, use (S2.11), and rearrange terms:

$$\begin{aligned} u\left(X(0)\right) &= u\left(X(T_x \wedge n)\right) + \int_0^{T_x \wedge n} f\left(X(s)\right)\mathrm{d}s \\ &\quad - \int_0^{T_x \wedge n} b\left(X(s)\right)\frac{\mathrm{d}}{\mathrm{d}x}u\left(X(s)\right)\mathrm{d}W(s) \\ &\quad - \int_0^{T_x \wedge n} \int_{\mathcal{Q}} u_{\Delta h}\left(X(s),q\right)\widehat{\mathcal{P}}\left(\mathrm{d}s,\mathrm{d}q;X(s),s\right). \end{aligned}$$

The expectation of the stochastic integrals are zero. Hence,

$$\lim_{n\to\infty}\mathbb{E}\left[u\left(X(0)\right)\mid X(0)=x\right] = \lim_{n\to\infty}\mathbb{E}\left[u\left(X(T_x\wedge n)\right) + \int_0^{T_x\wedge n} f\left(X(s)\right)\mathrm{d}s \,\bigg|\, X(0)=x\right]$$

$$u(x) = \mathbb{E}\left[\lim_{n\to\infty} u\left(X(T_x\wedge n)\right) + \int_0^{T_x\wedge n} f\left(X(s)\right)\mathrm{d}s \,\bigg|\, X(0)=x\right]$$

$$u(x) = \mathbb{E}\left[u\left(X(T_x)\right) + \int_0^{T_x} f\left(X(s)\right)\mathrm{d}s \,\bigg|\, X(0)=x\right]$$

$$u(x) = \mathbb{E}\left[g\left(X(T_x)\right) + \int_0^{T_x} f\left(X(s)\right)\mathrm{d}s \,\bigg|\, X(0)=x\right].$$

Ergo (S2.12) is justified. The limit and expectation can be interchanged as the argument of the expectation forms a uniformly integrable set (see [7]). □

S2-107   Similar to the previous theorem, this can also be extended to multiple dimensions with more
S2-108 bookkeeping. In multiple dimensions, this ordinary integro-differential equation would become a
S2-109 PIDE. See [7] for additional multi-dimensional special cases. The given reference considers problems
S2-110 without an integral term and provides assumptions that ensure the existence and uniqueness of the
S2-111 solution as well as the probabilistic representation. The proof provided here demonstrates that the
S2-112 integral term is not difficult to add in when the unique solution is assumed.

## S3-1 Supplementary Note 3: Additional Information on Main Text Examples

This note provides additional context and information for the example problems considered in the main text. The primary purpose of this section is to provide the equations and parameter values used along with any relevant discussion on building the particular discrete-time, discrete-space Markov chains associated with each example. Examples appear in the same order as the main text with particle transport problems appearing in Section S3.1 and heat equations on non-Euclidean geometries appearing in Section S3.2.

### S3.1 Particle Transport

In this section we will discuss further details of the particle transport examples in the main text. We first begin with some notation.

Assume that a particle occupies a single point with no mass and that intra-particle interactions can be ignored. The position and velocity of a particle is given by

$$\mathbf{r} = (x_1, x_2, x_3) \quad \text{and} \quad \mathbf{v} = v\Omega = v\left(\sin\theta\cos\phi, \sin\theta\sin\phi, \cos\theta\right),$$

respectively. The quantity $v$ is the particle speed. The angular density of particles at position $\mathbf{r}$ traveling in direction $\Omega$ with energy $E$ at time $t$ is denoted by $N(t, \mathbf{r}, \Omega, E)$.

A quantity of interest when considering particle transport is the angular flux density, given by

$$\Phi(t, \mathbf{r}, \Omega, E) = vN(t, \mathbf{r}, \Omega, E).$$

We will assume that:

- energy $E$ remains constant;

- we only care about either a single dimension or a projection into a single dimension, so that we may write $\mathbf{r} = x$ and $\Omega = \cos\theta$, or similar.

Under these assumptions, $\Phi := \Phi(t, x, \Omega)$. The angular flux density is then assumed to satisfy the Boltzmann particle transport equation given by

$$\frac{1}{v}\frac{\partial}{\partial t}\Phi(t, x, \Omega) + \Omega\frac{\partial}{\partial x}\Phi(t, x, \Omega) + \Sigma_t(x, \Omega)\Phi(t, x, \Omega) \\ = \int \Phi(t, x, \Omega')\,\sigma_s(x, \Omega' \to \Omega)\,\mathrm{d}\Omega' + R(t, x, \Omega), \quad (S3.1)$$

where

$$\Sigma_t(x, \Omega) = \Sigma_a(x, \Omega) + \Sigma_s(x, \Omega),$$

and $\Sigma_a$ and $\Sigma_s$ are functions representing the rates of particle absorption and scattering, and $R$ is a function representing a particle source [5]. The scattering term in the integral $\sigma_s$ is the scattering kernel and is related to the scattering rate function by

$$\Sigma_s(x, \Omega) = \int \sigma_s(x, \Omega \to \alpha)\,\mathrm{d}\alpha. \quad (S3.2)$$

S3-25    Note that the arrow indicates a transition in direction, and is not a limit. Given that a scattering
S3-26    event has occurred at position $x$, let $\eta$ represent the direction before scattering and $\alpha$ represent the
S3-27    direction of the particle after scattering. Then, the probability density function of $\alpha$ given $\eta$ and $x$
S3-28    is defined as

$$p(\alpha \,|\, \eta, x) = \frac{\sigma_s(x, \eta \to \alpha)}{\int \sigma_s(x, \eta \to \alpha')\,\mathrm{d}\alpha'} = \frac{\sigma_s(x, \eta \to \alpha)}{\Sigma_s(x, \eta)}. \tag{S3.3}$$

S3-29    Similarly, we can define the distribution of the reverse situation. That is, the probability density
S3-30    function of having came from the direction $\eta$ given that a scattering event occurred at position $x$
S3-31    and the direction after scattering was $\alpha$. Letting

$$S_\Sigma(x, \Omega) = \int \sigma_s(x, \eta \to \Omega)\mathrm{d}\eta, \tag{S3.4}$$

S3-32    we can write

$$p^*(\eta \,|\, \alpha, x) = \frac{\sigma_s(x, \eta \to \alpha)}{\int \sigma_s(x, \eta' \to \alpha)\,\mathrm{d}\eta'} = \frac{\sigma_s(x, \eta \to \alpha)}{S_\Sigma(x, \alpha)}. \tag{S3.5}$$

Note that when $p$ is symmetric, that is $p(a \,|\, b) = p(b \,|\, a)$ for all $a$ and $b$ then $\Sigma_s = S_\Sigma$ and $p = p^*$.
By using the above definitions and by assuming that $S_\Sigma > 0$, we can write

$$\int \Phi(t, x, \Omega')\sigma_s(x, \Omega' \to \Omega)\mathrm{d}\Omega' = S_\Sigma(x, \Omega) \int \left(\Phi(t, x, \Omega') - \Phi(t, x, \Omega)\right) p^*(\Omega' \,|\, \Omega, x)\,\mathrm{d}\Omega' \\ + S_\Sigma(x, \Omega)\Phi(t, x, \Omega). \tag{S3.6}$$

Now, by using (S3.6) and using the change of variables $\omega = \Omega' - \Omega$, the PIDE (S3.1) can be written as

$$\frac{\partial}{\partial t}\Phi(t, x, \Omega) = -v\Omega\frac{\partial}{\partial x}\Phi(t, x, \Omega) - v\left(\Sigma_t(x, \Omega) - S_\Sigma(x, \Omega)\right)\Phi(t, x, \Omega) + vR(t, x, \Omega) \\ + vS_\Sigma(x, \Omega) \int \left(\Phi(t, x, \Omega + \omega) - \Phi(t, x, \Omega)\right) p^*(\omega + \Omega \,|\, \Omega, x)\,\mathrm{d}\omega. \tag{S3.7}$$

S3-33    Note that the scattering and absorption related terms could also depend on time. We elected to
S3-34    omit that dependence here for compactness.

### S3-35  S3.1.1  Example 1: Simplified Transport

S3-36    The first example we consider is an angular flux density problem that only depends on direction
S3-37    $\Omega$ and not space $x$. Essentially we consider a hypothetical particle that has two states: state 1
S3-38    with $\Omega = 1$ and state 2 with $\Omega = -1$. The particle is subject to scattering and absorption events
S3-39    controlled by the constants $\Sigma_s$ and $\Sigma_a$ respectively. After scattering, the particle changes from
S3-40    state $i$ to state $j$ with probability $p_{ij}$. Note that $p_{ii}$ or $p_{jj}$ represents the probability that a particle
S3-41    does not change its state on a scattering event.

Since $\Phi$ does not depend on $x$ in this example, the parameter $v$ serves only to scale the absorption and scattering rates. Hence, we set $v = 1$ for clarity. Administering an initial condition, we seek

to solve the PIDE

$$\frac{\partial}{\partial t}\Phi(t,\Omega) = -(\Sigma_a + \Sigma_S - S_\Sigma(\Omega))\Phi(t,\Omega)$$
$$+ S_\Sigma(\Omega)\int(\Phi(t,\Omega+\omega) - \Phi(t,\Omega))p^*(\omega+\Omega|\Omega)\,\mathrm{d}\omega \quad \text{(S3.8)}$$
$$\Phi(0,\Omega) = g(\Omega) = \begin{cases} 5 & \text{if } \Omega = 1, \\ 3 & \text{if } \Omega = -1. \end{cases}$$

If we assume that $p_{ij} = 1/2$ for all $i,j$, then the analytic solution is given by

$$\Phi(t,\Omega) = \begin{cases} \frac{5}{2}\left(e^{-\Sigma_a t} + e^{-(\Sigma_a+\Sigma_s)t}\right) + \frac{3}{2}\left(e^{-\Sigma_a t} - e^{-(\Sigma_a+\Sigma_s)t}\right) & \text{if } \Omega = 1, \\ \frac{5}{2}\left(e^{-\Sigma_a t} - e^{-(\Sigma_a+\Sigma_s)t}\right) + \frac{3}{2}\left(e^{-\Sigma_a t} + e^{-(\Sigma_a+\Sigma_s)t}\right) & \text{if } \Omega = -1. \end{cases} \quad \text{(S3.9)}$$

For this example, we will further assume that $\Sigma_a = 0.5$ and $\Sigma_s = 5.0$.

Since $p_{ij} = 1/2$ for all $i$ and $j$, the distribution $p$ is symmetric and therefore $p = p^*$ and $S_\Sigma = \Sigma_s$. The probabilistic representation gives

$$\Phi(t,\Omega) = \mathbb{E}\left[e^{-\Sigma_a t}g(Y(t))\,\middle|\,Y(0) = \Omega\right], \quad \text{(S3.10)}$$
$$\mathrm{d}Y(t) = \omega_{Y(t)}\mathrm{d}P(t).$$

The stochastic process $Y(t)$ is effectively a proxy for our hypothetical particle. The particle retains its state (either 1 or $-1$) until the Poisson process $P(t)$ fires. Once the process fires, $Y(t)$ increases by the random change in direction $\omega_{Y(t)}$. We have included the subscript $Y(t)$ on $\omega$ to draw attention to the fact that the change in direction *depends* on direction before the Poisson process fires. We would need to base this random change in direction off of the reverse distribution $p^*$, meaning our proxy particle is moving in reverse when compared to the physical process. To tie this back to the original equation, the stochastic process is having us randomly select a previous direction $\Omega'$ given the current direction (after scattering) $\Omega$ every time the Poisson process fires. Luckily for this example, our forward distribuiton $p$ is symmetric – if we did not have $p_{ij} = 1/2$ for all $i,j$ then we would have to carefully draw our changes for the process $Y$.

Returning to the stochastic process, notice that it does not involve absorption. Rather, absorption is handled by the exponential term inside the expectation.

In order to develop the discrete-time Markov chain used to approximate the process $Y(t)$, we must first discuss the state space. For this problem, the state space is merely $\pm 1$, so we do not need to discretize by choosing some increment $\Delta\Omega$. The next step in creation of the Markov chain is selecting an appropriate time discretization $\Delta t$.

Note that for any $t$ and any $\Omega$, the parameter of the Poisson process $P(t)$ is

$$\int_t^{t+\Delta t} S_\Sigma \mathrm{d}u = S_\Sigma \Delta t = \Sigma_s \Delta t.$$

Ergo, our selection of $\Delta t$ is independent of both $t$ and $\Omega$. In particular, we will want to choose $\Delta t$ sufficiently small so that we can be reasonably sure that the Poisson process $P(t)$ will not fire more than one time in the time window $\Delta t$. Since the probability of no events occurring during the time window $\Delta t$ is

$$q_0 = e^{-\Sigma_s \Delta t},$$

and the probability of one event occurring is

$$q_1 = \Sigma_s \Delta t e^{-\Sigma_s \Delta t},$$

we want to choose $\Delta t$ so that

$$q_{>1} = 1 - q_0 - q_1$$

is sufficiently small, or less than 0.05. Given the value of $\Sigma_s$, a selection of $\Delta t = 0.01$ causes $q_{>1} \approx 0.001$.

For the construction of our Markov chain, we use $q_1$ for the probability that a Poisson event occurs and $1 - q_1$ for the probability that no Poisson event occurs. Technically, the stochastic differential equation representation for $Y(t)$ would have us calculate the probability for all increments in direction $\omega$ given the current direction and assign these probabilities to the states they would ultimately transition to. However, *for this problem* it is equivalent to instead change direction based on the given probabilities $p_{ij}$. Hence, we can define our Markov chain by the transition matrix

$$C = \begin{bmatrix} p_{11}q_1 + (1 - q_1) & p_{12}q_1 \\ p_{21}q_1 & p_{22}q_1 + (1 - q_1) \end{bmatrix}. \tag{S3.11}$$

This transition matrix was used to inform a random walk process on TrueNorth. As shown in **Fig. 3c** and **Fig. 3d** in the main text, 1000 and 10000 random walks per starting location were sampled, respectively. These were averaged according to (M.7) to produce the curves shown. The analytic solution is also plotted for comparison.

The low bit resolution of the PRNG for the stochastic configuration used for our implementation forced the transition probabilities to be quantized to an 8-bit resolution. For this problem, with our defined parameters

$$C \approx \begin{bmatrix} 0.976219264387482 & 0.0237807356125179 \\ 0.0237807356125179 & 0.976219264387482 \end{bmatrix}.$$

However, the actual transition probabilities used by the TrueNorth implementation are

$$C \approx \begin{bmatrix} 0.9765625 & 0.0234375 \\ 0.0234375 & 0.9765625 \end{bmatrix} = \frac{1}{256} \begin{bmatrix} 250 & 6 \\ 6 & 250 \end{bmatrix}.$$

The simulation was run for 5.5E+6 hardware ticks. This was to ensure at least 500 simulation steps were produced, equating to a real simulation time of 5 seconds. The images shown in **Fig. 3c-d** in the main text are zoomed to show $t \in [0, 2]$. Due to the random nature of the simulation, it is impossible to know, *a priori* exactly how many hardware ticks are needed in order to obtain exactly 500 simulation steps. Hence we must over estimate the run time and manually terminate.

<sub>S3-88</sub> ### S3.1.2  Example 2: Particle Angular Fluence

The second particle transport example we consider pertains to angular fluence, or time-integrated flux. Up to changes in units, the angular fluence problem is given by (S3.7), but with the time derivative set to zero and no dependence on time. For our simplified example, we will assume that $\Sigma_s$ is a constant and that $\Sigma_a$ is zero. That is, there are no absorption events. We additionally assume that after scattering events, the new direction of particles is uniform on $[-1, 1]$, and that we are concerned with $x \in [-1, 1]$ only. The uniform assumption again means that $p$ is symmetric and $p = p^*$ and $S_\Sigma = \Sigma_S$. Finally, we assume that the source term does not depend on direction of travel. Taken with an absorbing boundary condition, this gives the following problem:

$$0 = -v\Omega \frac{\partial}{\partial x} \Phi(x, \Omega) + vR(x)$$
$$+ v\Sigma_s \int (\Phi(x, \Omega + \omega) - \Phi(x, \Omega)) p^*(\omega + \Omega \,|\, \Omega) \, d\omega, \quad x \in (-1, 1), \Omega \in [-1, 1], \quad \text{(S3.12)}$$
$$\Phi(1, \Omega) = 0, \quad \text{if } \Omega < 0,$$
$$\Phi(-1, \Omega) = 0, \quad \text{if } \Omega > 0.$$

<sub>S3-89</sub> For this example, we take $v = 200$, $\Sigma_s = 0.15$, and

$$R(x) = \begin{cases} 0.015 & \text{if } |x| < 0.5 \\ 0 & \text{otherwise.} \end{cases}$$

There is no readily available analytic solution to this problem. The probabilistic solution is

$$\Phi(x, \Omega) = \mathbb{E}\left[\int_0^{T_{x,\Omega}} vR(X(u)) \, du \,\bigg|\, X(0) = x, Y(0) = \Omega\right],$$
$$dX(t) = -vY(t)dt, \quad \text{(S3.13)}$$
$$dY(t) = \omega_{Y(t)} dP(t),$$
$$T_{x,\Omega} = \inf\{t > 0 \,|\, X(t) \notin [-1, 1], X(0) = x, Y(0) = \Omega\}.$$

<sub>S3-90</sub> Once again the stochastic process provides a proxy for our particle. The position $X(t)$ decreases <sub>S3-91</sub> according to its velocity (speed $v$ times current travel direction $Y(t)$), and the current direction of <sub>S3-92</sub> travel $Y(t)$ is updated by the random increment $\omega_{Y(t)}$ every time the Poisson process fires. This <sub>S3-93</sub> proxy particle again behaves like the reverse of the actual particles. The negative sign on the $X(t)$ <sub>S3-94</sub> term means that the particle moves in the opposite of the current direction $Y(t)$. For $Y(t)$, the <sub>S3-95</sub> increment $\omega_{Y(t)}$ is chosen according to the distribution in the integral – given the current direction <sub>S3-96</sub> $\Omega$, a new direction $\Omega'$ is chosen such that $\Omega'$ could have *scattered to* the direction $\Omega$ after a scattering <sub>S3-97</sub> event.

<sub>S3-98</sub>   To create the Markov chain approximation, we will need to discretize the state space. This <sub>S3-99</sub> involves choosing a $\Delta x$ and $\Delta \Omega$ to create bins across the domain. For reasons that will become <sub>S3-100</sub> clear, we will wait to pick $\Delta x$ until after we have chosen $\Delta t$. To begin, we select $\Delta \Omega = 1/15$. This <sub>S3-101</sub> creates 30 possible locations for $Y(t)$ in $[-1, 1]$, corresponding to the midpoints of 30 bins. From left to right, these midpoint values begin with $-1 + \Delta\Omega/2$ and end at $1 - \Delta\Omega/2$, increasing by $\Delta\Omega$. We will write the midpoints of these bins as $\Omega_j = -1 + \Delta\Omega/2 + (j-1)\Delta\Omega$ for $j \in \{1, \ldots, 30\}$.

Next we will choose a value for $\Delta t$. For any time window $[t, t + \Delta t]$, the parameter for the Poisson process $P(t)$ is $v\Sigma_s \Delta t$. Using the same notation from the previous example for $q_0$, $q_1$, and $q_{>1}$, choosing $\Delta t = 0.01$ puts $q_{>1} \ll 0.05$.

Selecting both $\Delta t$ and $\Delta\Omega$ will help us choose a $\Delta x$ that will complement the problem. Note that in a single time window, a particle starting at some position $(x, \Omega)$ can only increment its position by $v\Omega\Delta t$. Once we discretized both direction and time, we have quantized the jumps the position can make based on the magnitude of the smallest nonzero direction. Since the smallest allowable (positive) direction is $\Delta\Omega/2$, this yields,

$$\Delta x = \frac{1}{2} v \Delta\Omega \Delta t. \tag{S3.14}$$

For our values of $v$, $\Delta\Omega$ and $\Delta t$, we have $\Delta x = 1/15$. This yields 30 possible spatial locations, corresponding to the midpoints of 30 bins in $[-1, 1]$. Again, we will denote these locations by $x_j$ where $x_j = -1 + \Delta x/2 + (j-1)\Delta x$, for $j \in \{1, \ldots, 30\}$. These divisions yield a state space of size $30 \times 30 = 900$.

Letting $(i, j)$ represent the location $(x_i, \Omega_j)$, we now seek to calculate the transition matrix

$$C^* = \left( c_{(i,j) \to (k,\ell)} \right),$$

where $c_{(i,j) \to (k,\ell)}$ represents the probability of transition from $(i, j)$ to $(k, \ell)$. We will assume some sort of ordering on the pairs $(i, j)$ so that $C^*$ is a $900 \times 900$ matrix. As implied in the previous discussion, the position can only transition to $x_i - v\Omega_j \Delta t$, a valid location by construction of $\Delta x$. As with the previous example, $p = p^*$. Therefore, it is equivalent to choose our new directions based on the final specified distributions of directions rather than calculating the appropriate conditional densities. In this example, we assumed directions after scattering are uniform. Therefore new directions in our discretized space are selected with probability $1/30$. Hence

$$c_{(i,j) \to (k,\ell)} = \begin{cases} (1 - q_1) + \frac{q_1}{30} & \text{if } x_k = x_i - v\Omega_j \Delta t \text{ and } j = \ell, \\ \frac{q_1}{30} & \text{if } x_k = x_i - v\Omega_j \Delta t \text{ and } j \neq \ell, \\ 0 & \text{otherwise.} \end{cases} \tag{S3.15}$$

This, however, does not define the entire transition matrix. We have not accounted for random walks that would travel out of the domain. This occurs when the transition from the state $(i, j)$ would create an $x$ value that no longer falls in $[-1, 1]$. To this end, we create an additional absorbing state, say **a** that corresponds to absorption. We must now define the column vector of probabilities

$$\mathbf{c_a} = \left( c_{(i,j) \to \mathbf{a}} \right).$$

The probability of transition to this state is simply 1 whenever the increment would force the walk to exit the domain. Hence,

$$c_{(i,j) \to \mathbf{a}} = \begin{cases} 1 & \text{if } |x_i - v\Omega_j \Delta t| > 1 \\ 0 & \text{otherwise.} \end{cases} \tag{S3.16}$$

S3-130 Finally, allowing $\overline{0}$ to represent a row vector of 900 zeros, the full $901 \times 901$ transition matrix for
S3-131 the Markov chain can be written as

$$C = \begin{pmatrix} C^* & \mathbf{c_a} \\ \overline{0} & 1 \end{pmatrix}. \tag{S3.17}$$

S3-132 This transition matrix was used to implement 6250 random walks per possible starting location on
S3-133 Loihi. Random walks are allowed to run until they arrive at the absorption state. The data from
S3-134 the collected random walks is averaged in the Monte Carlo sense via equation S3.13. We note that
S3-135 the averaging requires the use of a Riemann sum approximation to an integral. However, rather
S3-136 than keep information on every individual path, the Riemann sum term can be collapsed using
S3-137 cumulative densities. For more information on this technique, see [17]. The result is plotted as
S3-138 **Fig. 3g** in the main text.

S3-139 ## S3.2  Non-Euclidean Geometries

S3-140 The two previous examples utilize random walks on a domain that is not very complicated. The
S3-141 method does extend to more complicated domains, although there is some nuance in execution.
S3-142 Any time a diffusion coefficient exists ($a \neq 0$), the underlying SDE contains a Brownian motion
S3-143 term ($W(t)$). As detailed in **Supplementary Note 2**, this is a Brownian motion with respect
S3-144 to the appropriate probability space. If the problem were in $\mathbb{R}^d$, then $W(t)$ represents a standard
S3-145 $d$-dimensional Brownian motion.

S3-146 On the other hand, if the problem were on the surface of some admissible 3-dimensional shape,
S3-147 $W(t)$ is not a 3-dimensional Brownian motion, but rather a Brownian motion on the surface of
S3-148 the shape. For smooth shapes, this means that locally $W(t)$ is a 2-dimensional Brownian motion,
S3-149 however more complicated distributions can be defined (see [19] for a discussion on the von Mises-
S3-150 Fisher distribution, a distribution describing Brownian motion on the surface of the sphere).

S3-151 In this section we consider two examples monitoring heat transport on the surface of non-
S3-152 Euclidean objects. The first is a sphere; a smooth shape where locally the diffusion is a 2-
S3-153 dimensional Brownian motion. Here, transition probabilities are calculated by projecting to a
S3-154 tangent plane. The second is two spheres joined by a hexagonal prism. The shape is not smooth
S3-155 where the prism joins the spheres and along the spines of the prism. On the sphere, the same
S3-156 tangent plane approximation is used. On and near the prism, an unfolding argument is applied.
S3-157 See Figure S3.1 for a visualization of the sphere and barbell.

S3-158 ### S3.2.1  Heat Equation on the Surface of the Unit Sphere

Let $\mathbb{S}^2$ represent the unit sphere and consider the initial value heat equation:

$$\begin{aligned} \frac{\partial}{\partial t} u(t, \mathbf{x}) &= \alpha \nabla_{\mathbf{x}} u(t, \mathbf{x}), \qquad \mathbf{x} \in \mathbb{S}^2, \ t \in (0, \infty), \\ u(0, \mathbf{x}) &= g(\mathbf{x}). \end{aligned} \tag{S3.18}$$

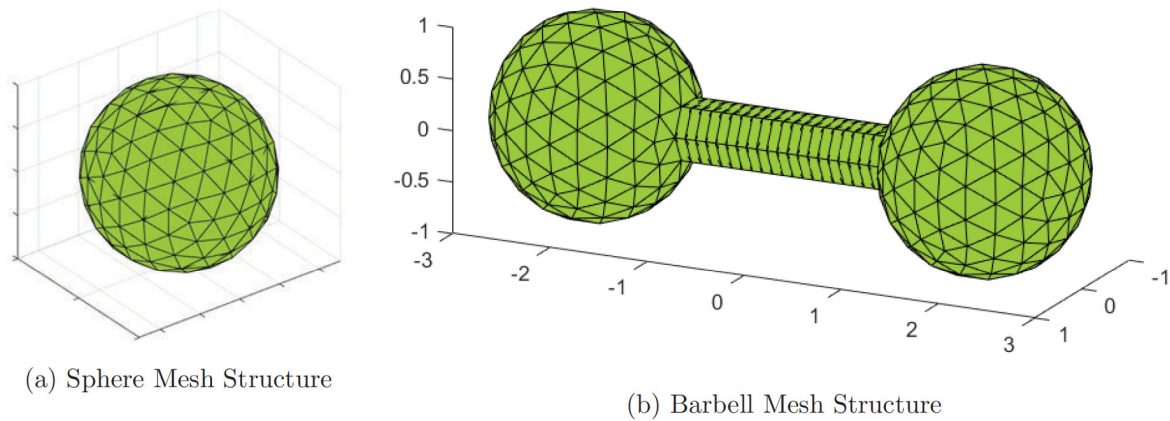

(a) Sphere Mesh Structure

(b) Barbell Mesh Structure

Figure S3.1: Visualization of mesh structure for heat transport examples. In the sphere (left), the center of each triangle represents a location in the mesh or an element of the state space. In the barbell (left), the center of each triangle or rectangle represents a location in the mesh.

With an appropriate $g$, the probabilistic solution is

$$u(t, \mathbf{x}) = \mathbb{E}\left[g\left(\mathbf{X}(t)\right) \mid \mathbf{X}(0) = \mathbf{x}\right]$$
$$d\mathbf{X}(t) = \sqrt{2\alpha}d\mathbf{W}(t), \quad \text{(S3.19)}$$

where $\mathbf{W}(t)$ is a Brownian motion on the surface of the sphere. Let $\mathcal{Y}_m^n$ represent the real part of the spherical harmonic of degree $m$ and order $n$. Suppose

$$g(\mathbf{x}) := g(\theta, \phi) = \mathcal{Y}_6^0(\theta, \phi) + \sqrt{\frac{14}{11}}\mathcal{Y}_6^5(\theta, \phi), \quad \text{(S3.20)}$$

where $(\theta, \phi)$ is the spherical coordinate for the vector $\mathbf{x}$ on the unit sphere with $\theta \in [0, \phi]$ and $\phi \in (-\pi, \pi]$. That is, $\theta$ measures the angle from the north pole and $\phi$ measures the angle from the prime meridian. Since the spherical harmonics are the eigenfunctions of the Laplacian in three dimensions, this initial condition admits a tidy analytic solution to (S3.18):

$$u(\mathbf{x}) := u(\theta, \phi) = e^{-42\alpha t}g(\theta, \phi). \quad \text{(S3.21)}$$

Let $\alpha = 1/42$ be given.

In order to perform our random walk approximation, we will first need to discretize the sphere. In the previous examples we took a uniform approach to discretization. In this example we will create roughly congruent triangles on the surface of the sphere. This is accomplished by calculating a geodesic dome structure.

Begin with an icosahdron with vertices on the surface of the unit sphere. An icosahedron is a polyhedron constructed with 20 equilateral triangles and 12 vertices. Our construction begins with two of these vertices equidistant from the north pole (and consequently two others equidistant from the south pole). At the midpoint of each triangle, a new line is drawn, creating four new equilateral triangles on each face of the icosahedron. Once more, on the midpoint of each triangle a new line is drawn creating four new smaller triangles. This yields 16 small equilateral triangles on each face of the original icosahedron for a total of 320 triangles. The vertices of these triangles are projected to the surface of the sphere (see Fig. S3.1a).

Our random walk on the sphere will traverse over the centroids of these triangles, projected to the surface of the sphere. We will allow the random walk to transition to any triangle neighbor that shares a vertex and back to itself. Since the original icosahedron had 12 vertices, this means that 60 triangles with have 12 possible transitions and the remaining 260 will have 13 possible transitions.

We must now select a time step size $\Delta t$ and compute transition probabilities for the triangles. We will ultimately use the tangent plane approximation method for this calculation. However, some may wonder why not use spherical coordinates since our initial condition and analytic solution utilize these coordinates. Translating (S3.18) into spherical coordinates yields

$$\frac{\partial}{\partial t} u(t,\theta,\phi) = \alpha \left( \frac{\partial^2}{\partial \theta^2} u(t,\theta,\phi) + \cot\theta \frac{\partial}{\partial \theta} u(t,\theta,\phi) + \csc^2\theta \frac{\partial^2}{\partial \phi^2} u(t,\theta,\phi) \right), \quad \text{(S3.22)}$$
$$u(0,\theta,\phi) = g(\theta,\phi).$$

This gives the probabilistic solution

$$u(t,\theta,\phi) = \mathbb{E}\left[ g(X(t),Y(t)) \mid X(0) = \theta, Y(0) = \phi \right]$$
$$dX(t) = \alpha \cot X(t) + \sqrt{2\alpha} dW_1(t) \quad \text{(S3.23)}$$
$$dY(t) = \sqrt{2\alpha \csc^2 X(t)} dW_2(t).$$

We can already see an issue with taking this approach – when $X(t)$ nears the north or south pole, the drift in latitude approaches infinity and the diffusion in longitude approaches infinity. We do not take this approach to avoid dealing with this singularity.

Order the 320 triangles in some fashion. Given a current triangle center $\mathbf{r}_i = (x_i, y_i, z_i)$, we project to a tangent plane using the gnomonic projection. This projects points to a tangent plane so that arcs along a great circle are projected onto straight line segments. We accomplish this by rotating the sphere so that $\mathbf{r}_i$ is at the north pole $(0,0,1)$ and then projecting all neighboring triangles to the $xy$ tangent plane centered at $(0,0)$. A standard two dimensional Brownian motion is used to approximate the probability of transition into each of these projected triangles. Under such a rotation and projection, the new coordinates $(x',y')$ of any point $(x,y,z)$ in terms of $(x_i,y_i,z_i)$ are given by

$$(x',y') = \begin{cases} \left(\frac{x}{z}, \frac{y}{z}\right) & \text{if } x_i^2 + y_i^2 = 0 \\ \left(\frac{z_i(x_i x + y_i y) - z(x_i^2 + y_i^2)}{\sqrt{x_i^2 + y_i^2}(x_i x + y_i y + z_i z)}, \frac{x_i y - y_i x}{\sqrt{x_i^2 + y_i^2}(x_i x + y_i y + z_i z)}\right) & \text{otherwise.} \end{cases} \quad \text{(S3.24)}$$

We note that in the construction we have detailed here, the north and south poles are always a vertex of a triangle and are never a center point. Therefore, the first assignment in (S3.24) is never used.

Suppose that $\mathbf{r}_i$ and $\mathbf{r}_j$ are centers of triangles that share a vertex. We use the SDE for $\mathbf{X}(t)$ (see (S3.19)) to inform our calculation of transition probability from $\mathbf{r}_i$ to $\mathbf{r}_j$. There is no drift term. Locally, $\mathbf{X}(t)$ is a two-dimensional Brownian motion. Letting $\mathbf{X}'$ represent the projection to the tangent plane, then $\mathbf{X}'(t)$ is Gaussian with mean $\mathbf{X}'(0)$ and covariance matrix given by

$$\Sigma_t = \begin{bmatrix} 2\alpha t & 0 \\ 0 & 2\alpha t \end{bmatrix}.$$

Applying Euler-Maruyama, $\mathbf{X}'(t+\Delta t)$ is Gaussian with mean $\mathbf{X}'(t)$ and covariance matrix given by

$$\Sigma_{\Delta t} = \begin{bmatrix} 2\alpha \Delta t & 0 \\ 0 & 2\alpha \Delta t \end{bmatrix}. \tag{S3.25}$$

Let $\tau_{ji}$ represent the triangle with center $\mathbf{r}_j$ in the tangent plane created by the gnomonic projection about the point $\mathbf{r}_i$. Let $f_\mathcal{G}(\rho, \mu, \Sigma)$ be the probability density function for the two dimensional Gaussian with mean $\mu$ and covariance matrix $\Sigma$ at the point $\rho$. Recalling that the projection of $\mathbf{r}_i$ is $(0,0)$, then we approximate the probability of transition from $\mathbf{r}_i$ to $\mathbf{r}_j$ by

$$p_{ij} \approx \int_{\tau_{ji}} f_\mathcal{G}(\rho, (0,0), \Sigma_{\Delta t}) \, d\rho, \tag{S3.26}$$

where $\Sigma_{\Delta t}$ is given by (S3.25). After selecting $\Delta t$, this integral can be evaluated numerically. We elected to use a Gaussian quadrature method. A selection of $\Delta t = 0.1$ ensures that the probability of transition to any triangle outside of those triangles that share a vertex with the starting location is less than 0.05. Departing from the previous examples, we added the missing probability for transition to the probability of not changing locations. That is, we added $1 - \sum_j p_{ij}$ to the probability $p_{ii}$. This is a simplifying choice we have made. Integration of transition to any other possible triangle via (S3.26) is not possible as the approximation is only valid locally and also because the gnomonic projection does not work for triangles in the opposite hemisphere.

Setting our transition matrix to

$$C = (p_{ij}), \tag{S3.27}$$

we implemented a graph for random walks on Loihi over the 320 possible locations. Using 3000 walkers total, the solution was calculated by changing the center points to spherical coordinates and averaging in the Monte Carlo sense via equation (S3.19). The simulation result displayed for various frames in time is given in **Fig. 4a** in the main text. Additionally, the norm of the difference in the Loihi calculated solution and the analytic solution over time for 1000 walkers/position is plotted in **Fig. 4b**.

### S3.2.2 Heat Transport on the Surface of a Barbell

Our final example is heat transport on a barbell shape. The barbell in consideration is two unit spheres joined by a hexagonal prism (see Fig. S3.1b). We created our barbell shape by starting with two unit spheres, centered on $\pm 2$. When performing the triangular mesh construction on the sphere as in the previous example, there will exist a left- or right-most vertex, closest to zero for each sphere. This vertex will belong to six triangles. We replace these triangles on either side with a hexagonal prism with side lengths equal to the length of the hexagon sides formed by the six triangles. Since the vertex we use to make this replacement comes from an initial division in the construction, this is a regular hexagon.

Let $\mathbb{B}$ be the surface of the two unit spheres joined by the described hexagonal prism. We are interested in solving

$$\frac{\partial}{\partial t} u(t, \mathbf{x}) = \alpha \nabla_x u(t, \mathbf{x}) - \kappa u(t, \mathbf{x}), \qquad \mathbf{x} \in \mathbb{B},\ t \in (0, \infty) \tag{S3.28}$$
$$u(0, \mathbf{x}) = g(\mathbf{x}).$$

The parameter $\kappa$ can be thought of as a rate of cooling. For this problem, we assume that $\alpha = 1/2$, $\kappa = 0.05$. Letting $\mathbf{x} = (x, y, z)$, we take

$$g(\mathbf{x}) = \begin{cases} 20 & y \geq 2.5, \\ 7 & 2.5 > y \geq 1, \\ 5 & 1 > y \geq 0, \\ 3 & 0 > y \geq -1, \\ 1 & -1 > y. \end{cases} \tag{S3.29}$$

The probabilistic solution is given by

$$u(t, \mathbf{x}) = \mathbb{E}\left[e^{-\kappa t} g(\mathbf{X}(t)) \,\middle|\, \mathbf{X}(0) = \mathbf{x}\right]$$
$$\mathrm{d}\mathbf{X}(t) = \sqrt{2\alpha}\,\mathrm{d}\mathbf{W}(t), \tag{S3.30}$$

where $\mathbf{W}(t)$ represents a Brownian motion on the surface of the barbell.

We discretize the spheres in the same manner from the previous example. We divide the hexagon into rectangles as follows. All the triangles on the sphere have roughly equal area by construction. We select one of the triangles with the smallest area; that is, one of the triangles that shares a vertex with the original vertices of the icosahedron. Using this area and the length of the edge of the hexagonal prism, we determine the ideal width of a rectangle to equal the area of this triangle. We then round this width based on the closest number of rectangles we can place along the prism.

From this construction, we get 314 triangles for each sphere and 120 rectangles in the prism, for a total of 748 locations in the state space. The states are taken to be the centroids of the triangles, projected to the surface of the sphere, and the centers of the 120 rectangles.

Probabilities of transition among triangles on the left and right spheres are handled like in the previous example. Triangles are again considered adjacent if they share a vertex. Transitions from a triangle on the sphere to a rectangle on the prism are only allowed if the rectangle shares a vertex with the triangle. Since there are six rectangles replacing six triangles on each sphere, and these rectangles are roughly equal in area to the triangles, we assign the probability of transition from a triangle to an adjacent rectangle to be the probability of transition from the triangle to the triangle that the rectangle is replacing. Again, this is not perfect. This is a choice we have made. The tangent plane projection does not work as well in these locations because the rectangles on the prism are almost perpendicular to the tangent plane.

Transitions among rectangles on the prism are allowed to other rectangles that share at least one vertex and back to the original rectangle. Transitions are calculated by unfolding the prism, setting the center of the rectangle equal to $(0,0)$, and calculating the probability of transitioning into the rectangles surrounding the current location. This probability is calculated via (S3.26), where the integration is performed over the appropriate rectangle rather than triangle.

Transitions from a rectangle on the prism to a triangle on the sphere are handled similarly. First, the triangle sharing a side with the rectangle is unfolded into the plane with the center of the rectangle occupying $(0,0)$. The probability of transition into this triangle is calculated via (S3.26). Now, depending on the rectangle (transitioning to left or right sphere), there are six additional triangles that share a vertex. These share the upper and lower vertices on one side of the rectangle, dependent on whether the rectangle is transitioning to the right or left sphere. There are three for the upper vertex and three for the lower vertex. We approximate the transition to one of these three upper triangles by calculating the probability of moving into the entire quarter plane diagonal from the upper vertex. This is accomplished by replacing the integration bounds in (S3.26) with the appropriate bounds for the quarter plane. This probability is divided by 3 and assigned to each of the three triangles sharing a vertex. Again, this is a simplifying choice as unfolding the three triangles is difficult in this scenario. This is repeated for the other three triangles touching the lower vertex.

Through this calculation, all possible transitions are calculated for locations that share a vertex. By selecting $\Delta t = 0.005$, we ensured that the probability of transition outside the allowable locations for each starting location in the state space was less than 0.05. As in the previous example, we add any missing probability to the probability of transitioning to the same location to ensure a total probability.

We use this to define a transition matrix. The graph for this random walk was implemented on a spiking net simulator as in [17]. Starting 1000 walkers on each location and averaging in the Monte Carlo sense via (S3.30), we calculated an approximate solution, plotted in **Fig. 4c** in the main text.